\documentclass[remotesensing,article,preprints,pdftex,moreauthors]{Definitions/mdpi}

\usepackage{amsmath,amssymb}

\usepackage{algorithm}        
\usepackage{algpseudocode}    

\usepackage{graphicx}
\usepackage{subfig}           

\usepackage{booktabs}         
\usepackage{array}            
\usepackage{tabularx}         
\usepackage{longtable}        
\usepackage{ltablex}          
\keepXColumns                 
\usepackage{siunitx}          
\newcolumntype{C}{>{$}c<{$}}
\newcolumntype{Y}{>{\centering\arraybackslash}X}

\firstpage{1} 
\pubvolume{1}
\issuenum{1}
\articlenumber{0}
\pubyear{2025}
\copyrightyear{2025}
\datereceived{ } 
\daterevised{ } 
\dateaccepted{ } 
\datepublished{ } 
\hreflink{https://doi.org/} 
\preto{\abstractkeywords}{\nolinenumbers} 

	\Title{TinyDef-DETR: A Transformer-Based Framework for Defect Detection in Transmission Lines from UAV Imagery}

	\TitleCitation{TinyDef-DETR: UAV-Based Transmission Lines Defect Detection}
	
	\Author{Feng Shen $^{1}$, Jiaming Cui $^{1}$, Wenqiang Li $^{1}$, and Shuai Zhou $^{1,2  *}$}
	
	\AuthorNames{Feng Shen, Jiaming Cui, Wenqiang Li, Shuai Zhou}
	
	\isAPAStyle{%
		\AuthorCitation{Shen, F., Cui, J., Li, W., Zhou, S.}
	}{%
		\isChicagoStyle{%
			\AuthorCitation{Shen, Feng, Jiaming Cui, Wenqiang Li and Shuai Zhou.}
		}{%
			\AuthorCitation{Shen, F.; Cui, J.; Li, W.; Zhou, S.}
		}
	}

	\address{%
		$^{1}$ \quad Harbin Institute of Technology, Harbin 150001, China;	fshen@hit.edu.cn(F.S.);hitcjm@stu.hit.edu.cn(J.C.);wenqli@stu.hit.edu.cn(W.L.);22B901046@stu.hit.edu.cn(S.Z.)\\
		$^{2}$ \quad Electric Power Research Institute, Yunnan Power Grid Co., Ltd., Kunming 650217,China; 22B901046@stu.hit.edu.cn(S.Z.)\\}
	
	\corres{Correspondence: 22B901046@stu.hit.edu.cn (S.Z.)}
	
	\abstract{
		Automated defect detection from UAV imagery of transmission lines is a challenging task due to the small size, ambiguity, and complex backgrounds of defects. This paper proposes TinyDef-DETR, a DETR-based framework designed to achieve accurate and efficient detection of transmission line defects from UAV-acquired images. The model integrates four major components: an edge-enhanced ResNet backbone to strengthen boundary-sensitive representations, a stride-free space-to-depth module to enable detail-preserving downsampling, a cross-stage dual-domain multi-scale attention mechanism to jointly model global context and local cues, and a Focaler-Wise-SIoU regression loss to improve the localization of small and difficult objects. Together, these designs effectively mitigate the limitations of conventional detectors. Extensive experiments on both public and real-world datasets demonstrate that TinyDef-DETR achieves superior detection performance and strong generalization capability, while maintaining modest computational overhead. The accuracy and efficiency of TinyDef-DETR make it a suitable method for UAV-based transmission line defect detection, particularly in scenarios involving small and ambiguous objects.
	}
	
	\keyword{Transmission lines;Defect detection;UAV imagery;Small Object Detection;}

	\addhighlights{yes}
	\renewcommand{\addhighlights}{%
		\noindent\textbf{\\What are the main findings?}
		\begin{itemize}[labelsep=2.5mm,topsep=-3pt]
			\item We propose \textbf{TinyDef-DETR}, a transformer-based detection framework specifically designed for UAV imagery in transmission line defect detection, integrating edge-enhanced feature extraction, stride-free space-to-depth downsampling, dual-domain multi-scale attention, and a Focaler-Wise-SIoU loss function.
			
		\end{itemize}
		
		\noindent\textbf{\\What is the implication of the main finding?}
		\begin{itemize}[labelsep=2.5mm,topsep=-3pt]
			\item The proposed framework enhances the reliability and scalability of UAV-based transmission line defect detection, enabling efficient defect analysis with low computational requirements and reducing missed detections in field deployments.
			\item The design principles of TinyDef-DETR demonstrate a certain level of generalizability in improving small-object detection performance, particularly in remote sensing and aerial vision applications.
		\end{itemize}
	}
	
\begin{document}
		
		
		\section{Introduction}
		
		Along with the sustained development of the Chinese economy, the mileage of transmission lines has increased rapidly\citep{yangReviewStateoftheArtPower2020}. As the core infrastructure for long-distance energy transmission, their reliable operation is essential for meeting the growing demand for electricity\citep{liuDataAnalysisVisual2020}. However, transmission lines are exposed to harsh outdoor environments\citep{chengAdInDETRAdaptingDetection2024}, where prolonged exposure significantly raises the risk of diverse faults—ranging from component damage to foreign object intrusions—collectively referred to as transmission line defects\citep{yangRecognitionBirdNests2023}.
		
		Unmanned aerial vehicles (UAVs) are increasingly employed to capture imagery of transmission lines for post-flight visual inspection and defect detection\cite{mohsanUnmannedAerialVehicles2022}. In such UAV-acquired imagery, defects often appear as small objects due to long shooting distances and the inherent resolution limitations of aerial imaging systems\cite{ahmedPowerTransmissionLine2024}. These small objects typically occupy regions smaller than 32×32 pixels, lacking clear contours and distinct textures, which severely hampers reliable recognition\citep{mirzaeiSmallObjectDetection2023}. The challenge is further exacerbated by complex backgrounds such as vegetation, towers, and shadows, where small defects are frequently obscured or confused with surrounding structures\citep{dengResearchProgressPower2024}.
		
		From a system perspective, the present work focuses on the \textbf{post-processing stage} of the inspection workflow, where UAVs execute routine patrols, capture high-resolution imagery along the transmission corridor, and subsequently upload these data to a ground-based server for automated analysis. In this paradigm, defect detection is conducted \textbf{offline as a post-processing task} rather than on-board the UAV, thus avoiding the stringent real-time and energy constraints associated with aerial platforms. This server-side processing setting enables the adoption of more expressive Transformer-based architectures and advanced optimization strategies while still meeting operational maintenance timelines. Meanwhile, the proposed model remains sufficiently lightweight to support potential migration to edge-computing platforms, achieving a practical balance between detection accuracy, robustness, and deployment flexibility.

		With the rapid advancement of smart grid development, transmission line defect detection has increasingly adopted deep learning-based object detection methods\cite{liCSSAdetRealTimeEndtoEnd2023}. Alongside this progress, numerous general-purpose detectors—such as YOLO\citep{redmonYouOnlyLook2016}, DINO\cite{zhangDINODETRImproved2022}, and RT-DETR\citep{zhaoDETRsBeatYOLOs2024}—have been developed. While these models achieve remarkable success in generic scenarios, most of them on the COCO dataset\cite{linMicrosoftCOCOCommon2014}, they face notable limitations in transmission line defect detection under complex conditions involving occlusion, background clutter, and variable lighting\cite{liuInsulatorDefectDetection2023a}. Common issues include missed detections, false positives, and false negatives, which substantially undermine the practical utility of UAV-acquired imagery in real-world transmission line maintenance\cite{dengResearchProgressPower2024a}. These shortcomings are consistently reported in previous studies and further confirmed by our preliminary experiments, highlighting the critical need for specialized detection frameworks capable of accurately identifying small-scale defects in UAV operational scenarios.
		
		In this article, we propose TinyDef-DETR, a novel transmission line defect detection network built upon the Transformer\citep{vaswaniAttentionAllYou2017} architecture, which aims to enhance detection accuracy, recall, and robustness for UAV-acquired imagery under complex inspection scenarios. The main contributions of this work are summarized as follows:
		\begin{enumerate}
			\item We construct an Edge Enhanced ResNet (EE-ResNet) by embedding an Edge-Enhanced Convolution (EEConv) module into the residual backbone. The EEConv integrates central, horizontal, and vertical difference convolutions to encode gradient priors, thereby improving sensitivity to fine-grained details and object boundaries. Through a re-parameterization strategy, the multi-branch structure of EEConv is fused into a single $3\times 3$ kernel at inference, incurring no additional computational overhead. By augmenting the ResNet backbone with such edge-aware feature extraction, EE-ResNet substantially strengthens the representation capability for small-scale defects in UAV inspection imagery.
			\item We introduce a stride-free downsampling module based on space-to-depth transformation, which reorganizes spatial information into the channel dimension using a fixed $2\times2$ partition without stride convolution. This design preserves pixel-level details while reducing spatial resolution, thereby improving recall for extremely small defects that are often missed by conventional strided convolution architectures.
			\item We propose a lightweight attention block with a dual-domain FSCA (Frequency-Spatial-Channel Attention) mechanism, integrated in a cross-stage partial manner. This design jointly enhances global semantic reasoning and local detail preservation, enabling the network to better capture anisotropic defect patterns such as missing bolts, loose clamps, and sagging wires.
			\item We develop a novel Focaler-Wise-SIoU loss that combines normalized IoU scaling, Scylla-IoU geometric penalties, and non-monotonic focal modulation. This adaptively emphasizes moderately difficult samples while suppressing trivial or noisy ones, thereby stabilizing training and improving localization accuracy for small-scale defect regions.
		\end{enumerate}

		\section{Related Works}
		\subsection{Applications of Transformer-Based Model in Power Line Defect Detection}
		Tian et al.\citep{tianAccurateEfficientInsulator2025} addressed the challenge of insulator defect detection in low-quality UAV imagery by introducing a patch-based diffusion model for image restoration and an optimized DETR with Spatial Information Interaction and Feature Convergence Modules, achieving superior detection accuracy of 95.8\% particularly for small defects, while the limitations lie in incomplete coverage of defect types, restricted generalization across diverse grid structures, and computational inefficiency in real-time applications, primarily due to dataset constraints and model complexity. Cheng and Liu successively advanced Transformer-based approaches for insulator defect detection by first enhancing DETR\citep{chengImageBasedDeepLearning2022} with refined attention and feature optimization to improve localization accuracy over the baseline, and later proposing AdIn-DETR\citep{chengAdInDETRAdaptingDetection2024} with lightweight feature fusion and task-specific adaptation for real-time end-to-end detection, achieving superior precision–efficiency trade-offs, though both methods remain limited by reduced robustness under complex backgrounds and unseen categories, largely due to dataset insufficiency and the inherent constraints of Transformer complexity and speed. Wang et al.\cite{wangEnhancingGridReliability2024} developed a Transformer-based detection framework for advanced insulator defect identification, integrating multi-scale feature enhancement and context-aware attention to improve recognition accuracy and robustness, achieving state-of-the-art reliability in complex grid environments, while the model still suffers from efficiency limitations and reduced adaptability to rare defect patterns, mainly due to computational overhead and the scarcity of diverse annotated datasets. Xie et al.\citep{xiePowerDETREndtoendPower2024} proposed Power-DETR, a Transformer-based end-to-end detection framework for power line defect components, in which contrastive denoising learning and a hybrid query selection strategy were introduced to enhance small-object recognition and detection robustness, achieving superior accuracy over baseline DETR, yet the method still faces challenges in computational efficiency and generalization to highly variable field conditions, mainly due to increased model complexity and limited dataset diversity. Yang and Wang \citep{yangRecognitionBirdNests2023} proposed a hybrid detection framework combining YOLOv5 with Transformer-based DETR to recognize bird nests on transmission lines from small-sample datasets, introducing transfer learning and data augmentation to mitigate limited training data, which achieved reliable detection performance under constrained scenarios, while its limitation lies in reduced scalability and robustness for large-scale deployments, mainly due to dataset scarcity and the trade-off between lightweight training and generalization capacity.
		
		\subsection{Joint Frequency–Spatial Representations for Small Object Detection
		}
		\label{sec:Review of Joint Frequency–Spatial}
		A growing body of work argues that purely spatial attention is insufficient for detecting small, weak, and cluttered targets, and instead advocates \emph{joint frequency–spatial} representations that explicitly preserve high-frequency cues while suppressing background aliasing and downsampling loss. Representative approaches inject wavelet- or spectrum-domain operators into DETR/YOLO backbones: for instance, replacing standard operators with wavelet-domain anti-aliasing and convolution improves fine-structure perception and multi-scale representation under real-time constraints \cite{shaoWTDETRWaveletenhancedDETR2025}. In infrared small-target detection, Haar wavelet decompositions have been used to derive high-/low-frequency masks that selectively retain fine details and mitigate over-smoothing, achieving state-of-the-art performance and illustrating an “frequency-guides-spatial” paradigm \cite{fengFADetFrequencyAwareDetection2025}; subsequent designs integrate frequency–spatial convolutions and cross-domain attention to highlight local high-frequency details while using low frequencies to complete global structure, yielding consistent gains with lightweight models \cite{xuThinkLocallyAct2025}. From the feature fusion perspective, frequency-gated attention strengthens high-frequency details that are easily drowned out by deep downsampling, and combining it with grouped spatial–channel transformers produces robust small-target descriptors \cite{liFSCNetFeatureSynthesis2025}; similarly, explicitly separating and re-fusing spatial and frequency channels in hierarchical transformers alleviates the loss of detail caused by aggressive subsampling and semantics-dominant deep layers \cite{wangSmallObjectDetection2025}. Importantly, these studies also caution that frequency interaction is not “the more the better”: without proper masking or dynamics modeling, it may amplify boundary noise; coupling frequency-interaction attention with stabilizing mechanisms has been shown to suppress irrelevant high-frequency artifacts and emphasize targets nearly indistinguishable from background \cite{chenFreqODEsFrequencyNeural2024}. Beyond wavelets, discrete cosine transform (DCT) features, aligned and fused with spatial cues, have improved AP for small objects in remote sensing, indicating that cosine-spectrum bases offer complementary discriminative power for tiny structures \cite{luoFrequencySpectrumFeatures2024}. Earlier frequency/multi-resolution methods (e.g., NSCT/Contourlet families) have also informed background estimation and low-frequency isolation: isolating low-frequency components reduces sensor and clutter noise while high-frequency components recover target details, providing a principled route to boost the small-target signal-to-noise ratio \cite{liMitigateTargetLevelInsensitivity2024,liuInfraredAerialSmall2018}. In UAV scenarios emphasizing on-board efficiency, integrating wavelet convolutions and lightweight attention into real-time DETR variants demonstrates stable gains on diminutive, high-altitude objects and underscores the efficacy of “frequency-aware convolutions + transformers” under embedded constraints \cite{guoAUHFDETRLightweightTransformer31}. 
		
		In summary, the literature converges on three lessons. First, frequency decomposition should be paired with spatial alignment or masking to suppress noisy high-frequency components \cite{fengFADetFrequencyAwareDetection2025,chenFreqODEsFrequencyNeural2024}. Second, coupling with multi-scale or large-receptive-field modules is crucial to balance local details and global context \cite{shaoWTDETRWaveletenhancedDETR2025,xuThinkLocallyAct2025}. Finally, the practical impact depends on lightweight implementations suitable for UAV platforms\cite{guoAUHFDETRLightweightTransformer31}.	
		\section{Materials and Methods}
		\subsection{Overview}
		
		In this study, we employ RT-DETR-r18\cite{zhaoDETRsBeatYOLOs2024} as the baseline detection framework owing to its superior accuracy and robustness in complex scenarios. Compared with traditional DETR models, RT-DETR exhibits significantly faster training convergence and improved deployment efficiency, making it more suitable for real-time and resource-constrained applications such as UAV-based inspection tasks \cite{suvittawatAdvancesAircraftSkin2025}.
		
		RT-DETR incorporates an enhanced encoder–decoder architecture together with refined query interaction mechanisms, while for the backbone network, RT-DETR-r18 adopts a CNN-based design, namely the popular ResNet-18\cite{heDeepResidualLearning2016}. From the perspective of real-time performance, employing a CNN architecture for feature extraction further enhances the real-time capability of DETR-like models, thereby improving their practical applicability. As a result, RT-DETR not only accelerates convergence and improves detection precision, but also preserves the global context modeling capability of Transformer-based frameworks\citep{vaswaniAttentionAllYou2017}.In addition, its lightweight optimization strategies—specifically, the design choice to apply the encoder only to high-level semantic features (S5) with relatively small spatial dimensions—effectively reduce resource consumption. This balances performance and speed by significantly decreasing computational overhead and accelerating inference without compromising accuracy, thereby making RT-DETR particularly suitable for post-flight UAV imagery analysis, where large volumes of data must be processed efficiently under computational and energy constraints\citep{tianAccurateEfficientInsulator2025}.
		
		By contrast, YOLO-based detectors, though widely favored for their real-time efficiency and ease of deployment\citep{vijayakumarYOLObasedObjectDetection2024}, tend to degrade in performance when handling small, occluded, or densely distributed objects, owing to their anchor-dependent design and limited global receptive field. This aligns with the findings of Ismail et al.\citep{ismailComparativeAnalysisDeep2024}, who reported that while YOLO excels in real-time scenarios, it struggles with small-object detection and occlusion, whereas DETR achieves higher accuracy in complex environments at the cost of greater computational demand. Our preliminary experiments further compared RT-DETR with representative YOLO series models, including YOLOv3\citep{redmonYOLOv3IncrementalImprovement2018}, YOLOv7\citep{wangYOLOv7TrainableBagofFreebies2023}, and YOLO 11\citep{khanamYOLOv11OverviewKey2024}, and showed that RT-DETR consistently outperformed most of them in overall performance. In cases where certain YOLO variants (e.g., YOLO 11m) surpassed RT-DETR, their gains were typically accompanied by substantially larger parameter sizes, thereby underscoring RT-DETR’s advantageous balance between model complexity and detection accuracy.
		
		Taken together, the consistent advantages of RT-DETR over both traditional DETR models and representative YOLO series substantiate its adoption as the baseline framework in this study.
		
\begin{figure}[h]
			\centering
			\includegraphics[width=1\textwidth]{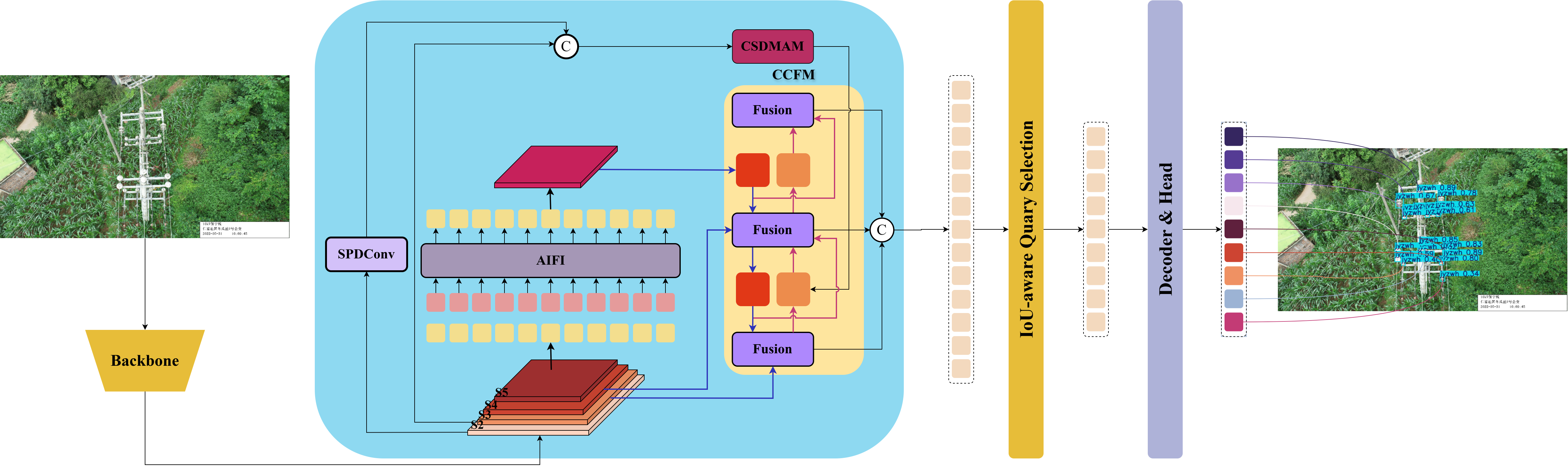}
			\caption{Overall framework of TinyDef-DETR}
			\label{TinyDefDETR}
		\end{figure}

		Building upon this strong foundation, we propose TinyDef-DETR as a dedicated solution for small object defect detection in UAV-based transmission line inspection. The overall architecture is illustrated in Fig.\ref{TinyDefDETR}. 
		
		First, the backbone adopts the Edge-Enhanced ResNet (EE-ResNet) structure to strengthen low-level detail preservation. Second, a compact Space-to-Depth Convolution (SPD) module is inserted before each downsampling layer to reorganize spatial features in a manner that preserves critical spatial information while reducing resolution, thereby improving the recall rate of extremely small objects. Third, a lightweight Cross-Stage Dual-Domain Multi-Scale Attention Module (CSDMAM) is embedded into the backbone network to enhance directional sensitivity and fine-grained structural modeling, effectively capturing anisotropic features of typical defects such as missing bolts, loose clamps, or sagging wires. Finally, a Focaler-Wise-SIoU loss function is employed to jointly optimize classification and localization. This loss function adaptively balances easy and hard samples while providing more accurate bounding box regression through shape-aware optimization, which further improves detection robustness for small and ambiguous objects.
		Detailed descriptions of the proposed modules can be found in Sections~\ref{sec:edge_resnet} to ~\ref{sec:loss}.
		
		\subsection{Edge Enhanced ResNet}
		\label{sec:edge_resnet}
		
		ResNet\citep{heDeepResidualLearning2016} represents a milestone in deep learning for computer vision, introducing residual learning that enables the effective training of very deep networks. Owing to its simple yet modular design, it has become a widely adopted backbone across a broad spectrum of vision tasks. Nevertheless, when applied to UAV-based visual inspection for transmission line defect detection, native ResNet architectures, as verified in preliminary experiments, exhibit limited capability in capturing sufficiently fine-grained features, thereby constraining their effectiveness in identifying small and subtle defect objects, particularly in challenging transmission line inspection environments with interference from vegetation, man-made structures, and adjacent electrical installations. 
		
		Inspection of the heatmaps exported from ResNet further reveals that they lack clear edges and fine contours, which exacerbates the difficulty of accurately localizing subtle defects.
		Unlike conventional convolution, where the high performance of CNN-based edge feature extraction typically relies on large pretrained backbones—resulting in high memory consumption and energy cost, which contradicts the real-time requirements of UAV-based transmission line defect detection tasks—difference convolution introduces differential operations to enhance sensitivity to gradient variations and fine-grained structural details, thereby improving the extraction of local details and object boundaries\cite{wangLCSCUAVNetHighPrecisionLightweight2025}.
		
\begin{figure}[htb]
			\centering
			\subfloat[EEConv (training)\label{fig:sub_a}]{
				\includegraphics[height=0.28\textwidth]{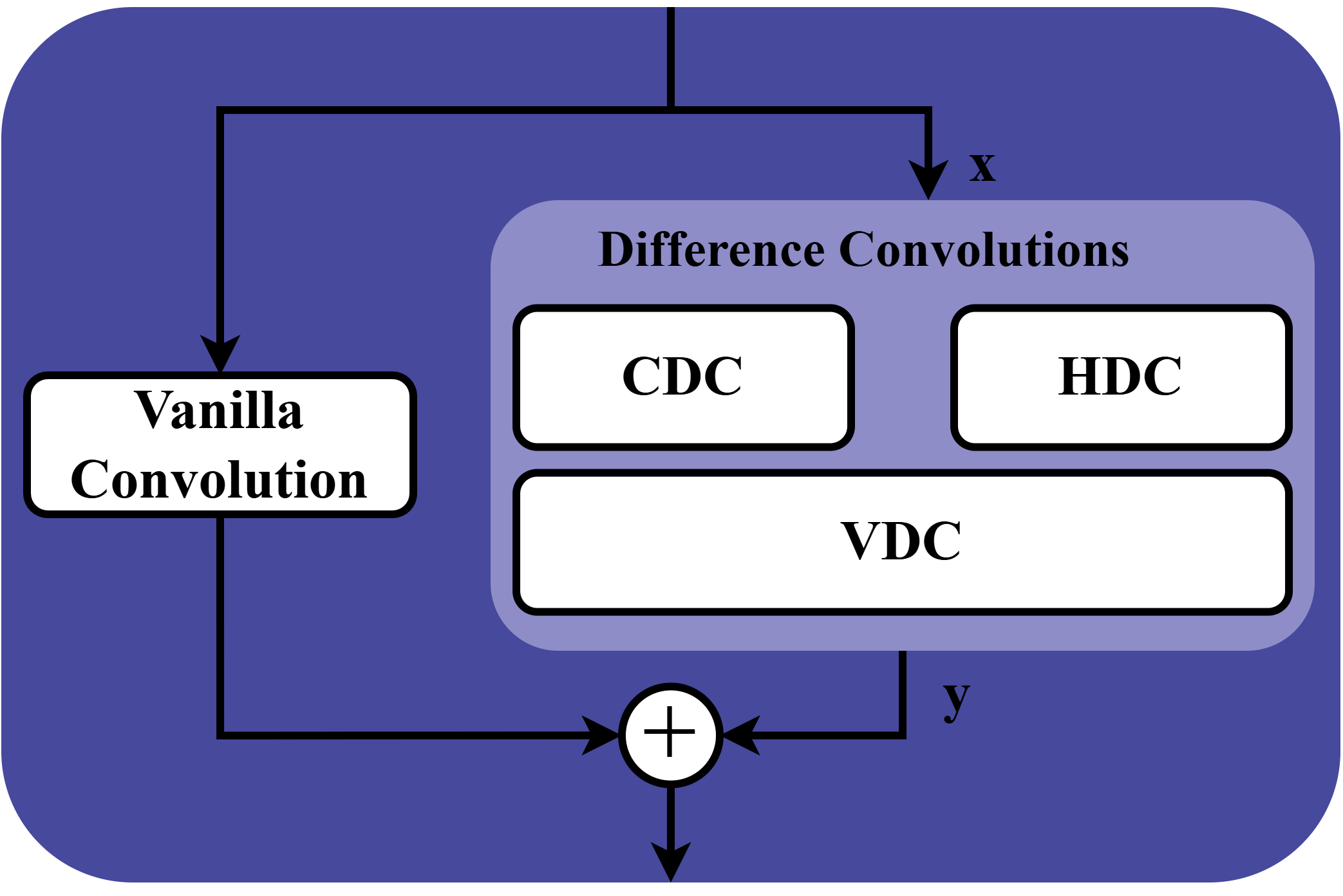}
			}\hfill
			\subfloat[EEConv (inference)\label{fig:sub_b}]{
				\includegraphics[height=0.28\textwidth]{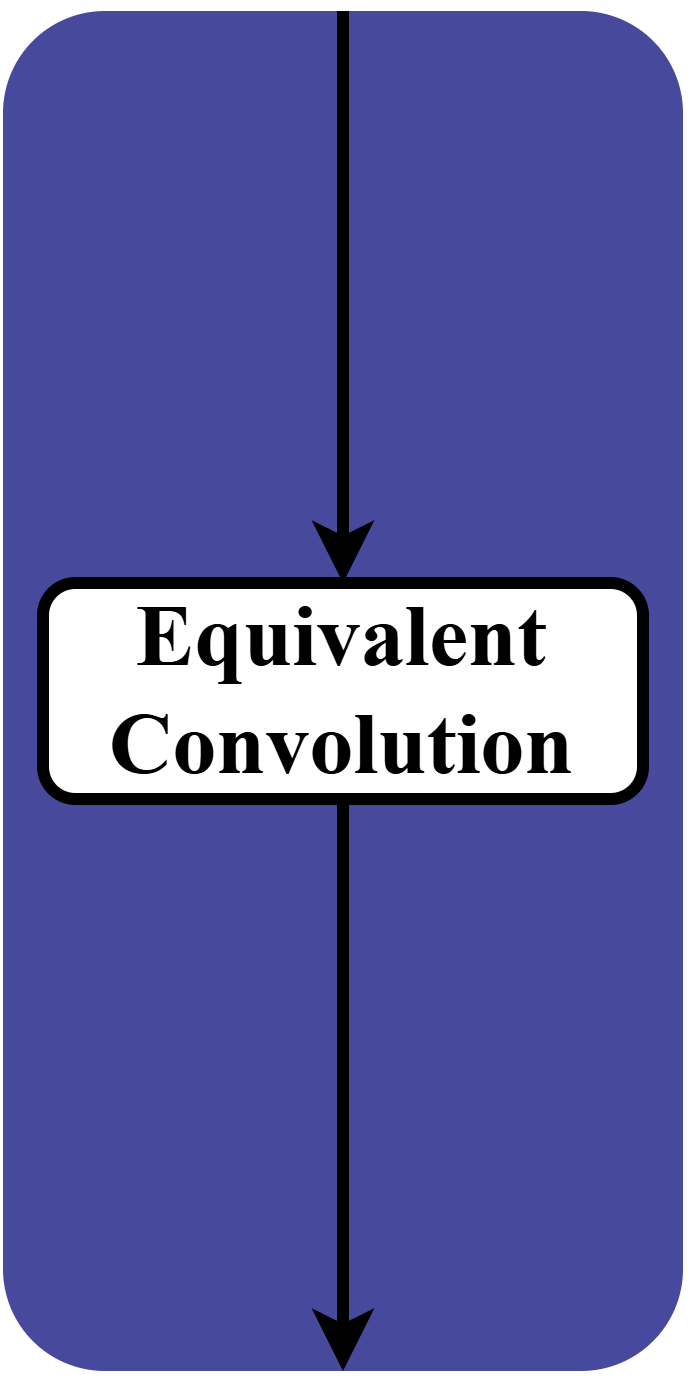}
			}\hfill
			\subfloat[EEBlock\label{fig:sub_c}]{
				\includegraphics[height=0.28\textwidth]{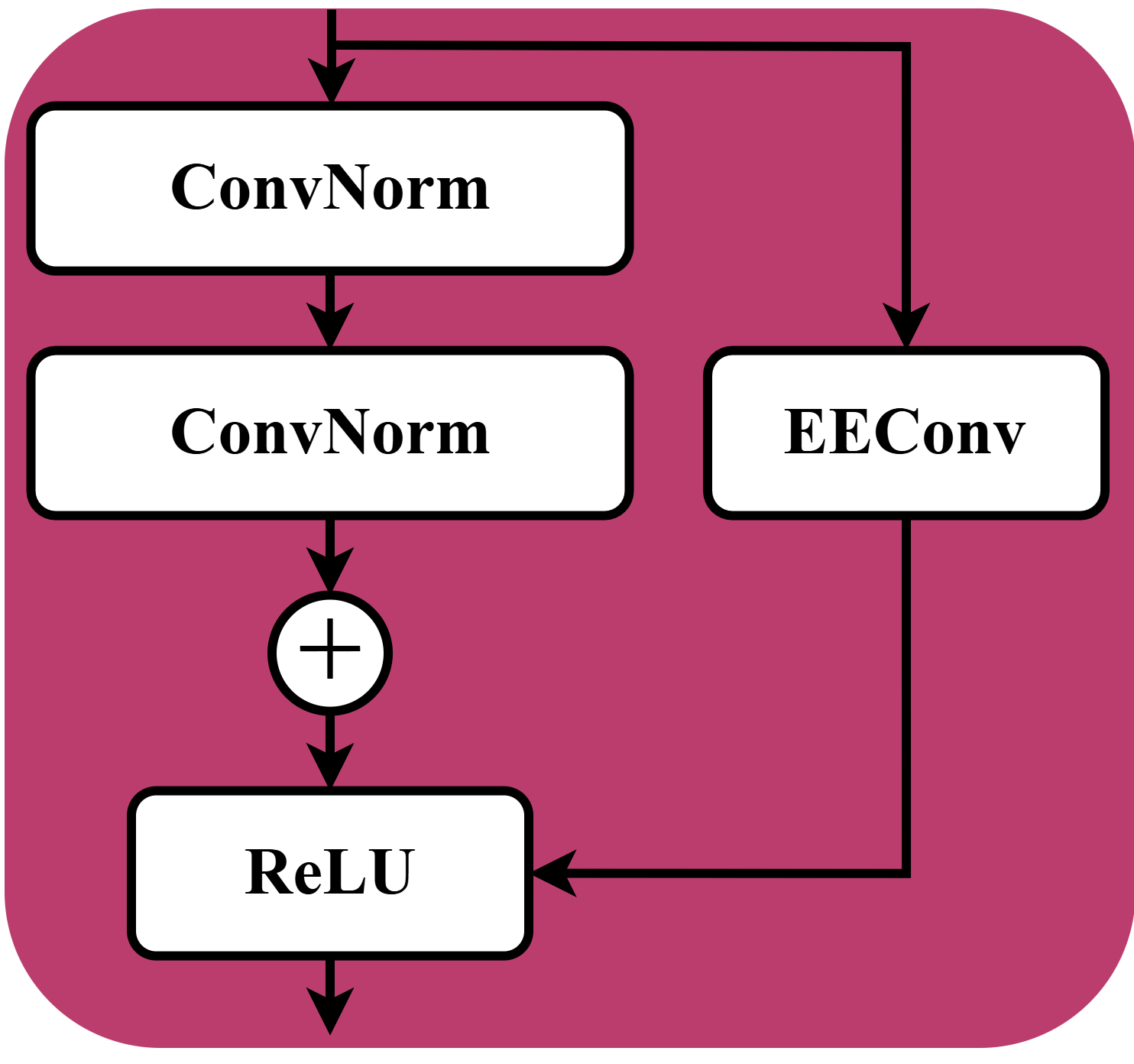}
			}
			\vspace{-6pt}
			\caption{The architecture of EEBlock (Edge Enhanced Block)}
			\label{fig:EEBlock}
		\end{figure}

		To address the aforementioned problem, we propose an EEConv (edge enhancement convolution). Building on this idea, EEConv explicitly incorporates three difference convolution operators—central difference convolution (CDC), horizontal difference convolution (HDC), and vertical difference convolution (VDC)—to embed gradient priors into the feature extraction process. The overall structure of EEConv is illustrated in Fig.~\ref{fig:sub_a}.To ensure deployment efficiency, EEConv adopts a re-parameterization strategy that fuses its multi-branch convolutions into a single equivalent 3×3 kernel during inference. Once fused, the auxiliary branches are discarded, leaving only a lightweight standard convolution. This allows the module to preserve the expressive power gained from multi-branch training while maintaining the same computational cost as a vanilla convolution.During deployment, the structure of EEConv becomes as shown in Fig.~\ref{fig:sub_b}.
		
		\vspace{-18pt}
		
		\vspace{-6pt}
		
		Formally, let the four convolutional branches be denoted as
		$(W_k, b_k),\; k=1,2,3,4$, corresponding to CDC, HDC, VDC, and the Vanilla convolution, respectively.
		During training, their weights and biases are linearly aggregated to obtain an equivalent $3\times3$ kernel and bias:
		\[
		\begin{aligned}
			W_{\Sigma} &= \sum_{k=1}^{4} W_k,\\
			b_{\Sigma} &= \sum_{k=1}^{4} b_k.
		\end{aligned}
		\]
		
		The intermediate feature map is then obtained as
		\[
		z = \text{Conv}_{3\times 3}(x; W_{\Sigma}, b_{\Sigma}),
		\]
		followed by batch normalization and a non-linear activation $\phi(\cdot)$:
		\[
		y = \phi\!\left( \gamma \cdot \frac{z - \mu}{\sqrt{\sigma^2 + \epsilon}} + \beta \right).
		\]
		
		For inference, the BN parameters are absorbed into the convolution kernel, yielding per-channel scaling:
		\[
		\alpha_c = \frac{\gamma_c}{\sqrt{\sigma_c^2 + \epsilon}}, 
		\]
		and the final equivalent kernel and bias are computed as
		\[
		\begin{aligned}
			W_c^{(\text{final})} &= \alpha_c \cdot W_{\Sigma,c}, \\
			b_c^{(\text{final})} &= \alpha_c \cdot (b_{\Sigma,c} - \mu_c) + \beta_c .
		\end{aligned}
		\]
		
		Thus, during deployment, EEConv degenerates into a single $3 \times 3$ convolution followed by activation:
		\[
		y = \phi\!\left( \text{Conv}_{3\times 3}(x; W^{(\text{final})}, b^{(\text{final})}) \right),
		\]
		which guarantees equivalence with the training-time formulation while ensuring zero additional inference overhead.
		
		To integrate this into our framework, we replace the second $3 \times 3$ convolution in the residual branch of standard ResNet basic blocks with EEConv, thereby preserving the backbone’s modular design while substantially improving sensitivity to fine-grained structures. This design choice is motivated by the observation that the second $3 \times 3$ convolution in each residual block primarily functions as a feature refiner, responsible for aggregating local structural details after the first convolution has extracted preliminary features. In this way, we obtain the Edge-Enhanced Block (EEBlock), whose structure is illustrated in Fig.~\ref{fig:sub_c}.
		
		By substituting the refining convolution with EEConv, we embed directional gradient priors precisely at the stage where fine-grained information is consolidated, thus maximizing the impact of EEConv on enhancing edge and detail representation. Meanwhile, the first $3 \times 3$ convolution and the shortcut connection remain unchanged, ensuring full compatibility with the original ResNet architecture and preserving its residual learning mechanism.
		
\begin{figure}[h]
			\centering
			\includegraphics[width=1\textwidth]{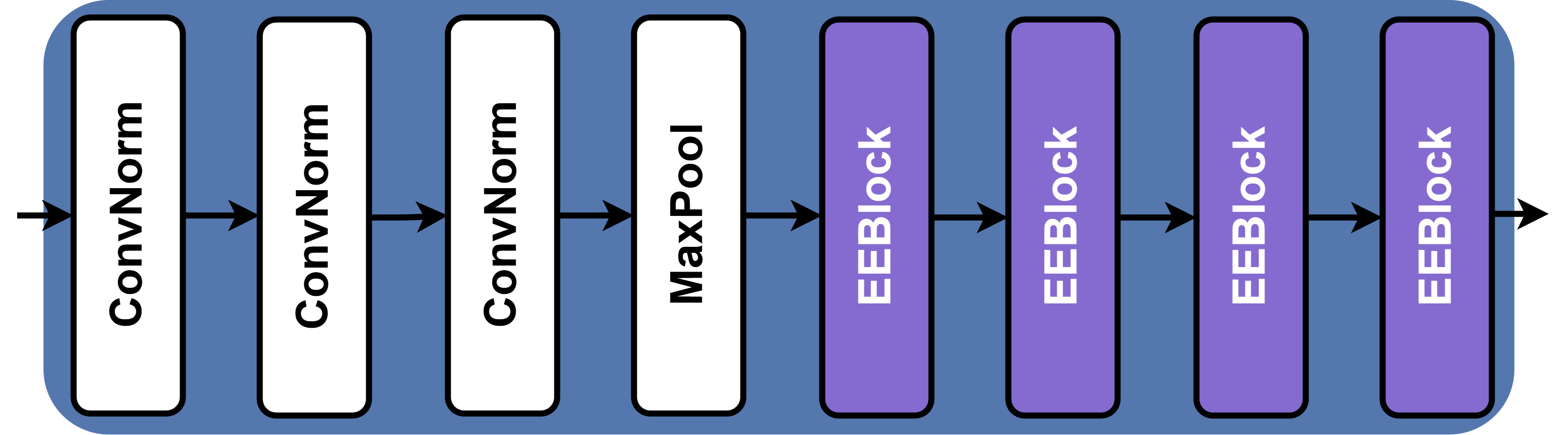}
			\caption{Overall framework of EE-ResNet}
			\label{EE-ResNet}
		\end{figure}

		Integrating EEBlocks throughout the backbone yields the Edge-Enhanced ResNet (EE-ResNet), which significantly improves sensitivity to subtle structural details while maintaining computational efficiency. EE-ResNet provides a robust feature extraction foundation, offering stronger boundary-aware representations and improved resilience against background interference. This enhanced backbone design directly addresses the core challenge of detecting small-scale and fine-grained defects in transmission line imagery, thereby serving as a crucial cornerstone for the subsequent TinyDef-DETR framework. The overall architecture of the proposed EE-ResNet is illustrated in Fig.~\ref{EE-ResNet}.
		
		\subsection{Space-to-depth Convolution}
		\label{sec:spdconv}
		\subsubsection{Principle of SPD(Space-to-depth Convolution)}
		Given the nature of transmission line defect detection—where the dataset is predominantly composed of small objects—preserving fine-grained information, maintaining balanced feature weighting, and performing effective feature representation learning contribute to improving both accuracy and recall \citep{liuDataAnalysisVisual2020}.
		
		During an extensive review of the literature, we came across the work of Sunkara and Luo \citep{10.1007/978-3-031-26409-2_27}, who pointed out that conventional CNN architectures exhibit an inherent limitation, namely the use of strided convolution. They emphasized that such designs inevitably lead to the loss of fine-grained information and result in less effective feature representations, which is particularly detrimental for low-resolution images and small object detection. This explains why conventional CNN-based approaches often fail to meet the requirements of small object detection, where the preservation of subtle spatial details is crucial. 
		
		Inspired by their findings, we incorporate the SPD(Space-to-depth Convolution) module into our proposed TinyDef-DETR framework. SPD adopts a stride-free downsampling strategy that redistributes pixels along the channel axis instead of discarding them, thereby preserving pixel-level information while reducing spatial resolution. In this way, TinyDef-DETR achieves more effective feature representations, improving both accuracy and recall for transmission line defect detection.
		
		The SPD operation is defined on an input feature map
		\[
		X \in \mathbb{R}^{S \times S \times C},
		\]
		where $S$ denotes the spatial resolution and $C$ is the channel dimension. 
		The goal is to downsample the spatial resolution by a factor of $\text{scale}$ while preserving all pixel-level information. 
		Unlike conventional strided convolution that inevitably discards part of the input, SPD achieves lossless downsampling by spatial-to-channel rearrangement. 
		Formally, the output feature map is given by
		\[
		X' \in \mathbb{R}^{\tfrac{S}{\text{scale}} \times \tfrac{S}{\text{scale}} \times (C \cdot \text{scale}^2)} .
		\]
		
		This transformation can be expressed as a structured partitioning of $X$. 
		Specifically, for each spatial position $(u,v)$ in $X'$, its corresponding feature vector is constructed by concatenating the values of $X$ sampled from the $\text{scale} \times \text{scale}$ local neighborhood in the original feature map:
		\[
		X'(u,v,:) 
		= \bigoplus_{i=0}^{\text{scale}-1} \, \bigoplus_{j=0}^{\text{scale}-1} 
		X(u \cdot \text{scale} + i, \, v \cdot \text{scale} + j, :)
		\]
		where $\bigoplus$ denotes concatenation along the channel dimension. 
		In this way, all pixels from $X$ are preserved in $X'$, but redistributed from the spatial dimensions into the channel dimension.
		
		After the SPD transformation, the spatial resolution of the feature map is reduced by a factor of $\text{scale}$, while all pixel-level information is preserved through spatial-to-channel rearrangement. 
		Consequently, the number of channels increases by a factor of $\text{scale}^2$. 
		To further process this rearranged representation, a subsequent convolution layer is applied, which can be formally expressed as
		\[
		X'' = f_{\text{Conv}}\bigl(X'; \mathbf{W}, \mathbf{b}\bigr), \quad 
		X'' \in \mathbb{R}^{\tfrac{S}{\text{scale}} \times \tfrac{S}{\text{scale}} \times C_2},
		\]
		where $\mathbf{W}$ and $\mathbf{b}$ denote the convolutional kernel weights and bias, respectively. 
		This convolution simultaneously fuses information across the expanded channel dimension and adjusts it to the object dimension $C_2$. 
		In this way, the operation not only regulates the channel dimensionality but also enhances local feature representation through the structured partitioning.
		
\begin{figure}[H]
			\centering
			\includegraphics[width=1\textwidth]{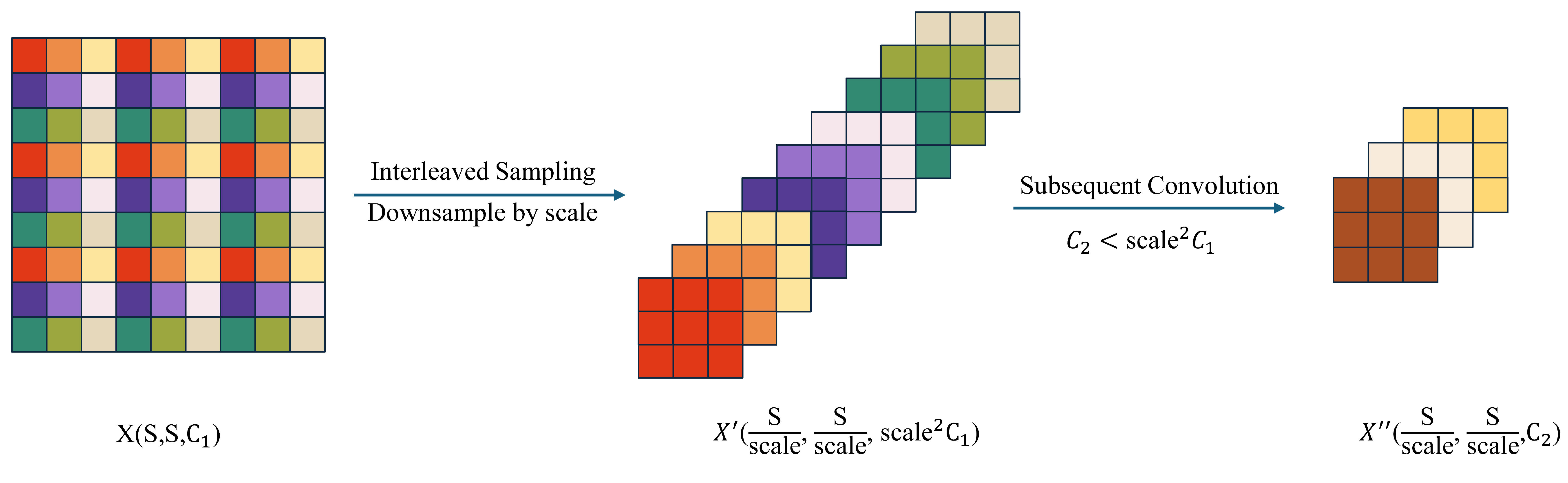}
			\caption{Illustration of SPDConv (Space-to-Depth Convolution)}
			\label{fig:spdconv}
		\end{figure}

		An illustrative example of the SPDConv operation is presented in Fig.\ref{fig:spdconv}.
		
		\subsubsection{Choice of the Downsampling Factor}
		A critical design decision in the SPD module is the choice of the downsampling factor \emph{scale}.
		
		\textbf{Constraint of Small Object Size.} In our dataset, small objects including polluted insulator, missing tie wire, and bird nest frequently span only $8 \sim 32$ pixels in the original image. If $\text{scale}=4$ (or larger) were adopted, the smallest objects would be reduced to merely $2 \sim 8$ pixels at the first feature level. After further downsampling by the backbone and feature pyramid, these objects would rapidly degenerate into near-point responses, severely hindering recall. By contrast, $\text{scale}=2$ preserves the effective resolution of small objects at the $4 \sim 16$ pixel level, which can still be robustly captured and aggregated by local $3 \times 3$ operators.
		
		\textbf{Compatibility with Detector Stride Hierarchy.} Unlike many mainstream detectors that organize feature pyramids with strides $\{4, 8, 16, \ldots\}$, our Edge Enhanced ResNet employs stage-wise downsampling with stride $=2$. Consequently, setting SPD $\text{scale}=2$ is equivalent to a lossless stride-2 operation, perfectly aligned with the subsequent stride-2 downsampling in the backbone/pyramid, preserving high-resolution shallow features crucial for small objects. In contrast, starting with $\text{scale}=4$ effectively performs a stride-4 rearrangement at the outset, over-compressing shallow maps and weakening the small-object modeling capacity of early pyramid levels.
		
		\textbf{Anti-Aliasing and Information Fidelity.} With $\text{scale}=2$, the neighborhood is completely mapped into the channel dimension, ensuring that no samples are discarded. Although $\text{scale} \geq 4$ also performs rearrangement, the resulting features must be processed at a much lower spatial resolution, reducing the discriminability of local topology and diminishing the effective anti-aliasing capability.
		
		\textbf{Efficiency in Computation and Memory.} After SPD, the channel dimension increases by a factor of $\text{scale}^2$. When $\text{scale}=2$, channels are quadrupled while spatial dimensions are reduced by one quarter. As a result, the computational cost of the following $3 \times 3$ convolution remains comparable to that of a standard stride-2 convolution, but without discarding information. In contrast, $\text{scale}=4$ would increase the channel dimension by sixteen while reducing spatial size by one sixteenth, leading to heavier memory and bandwidth requirements and potentially unstable training.
		
		Taken together, these factors demonstrate that setting $\text{scale}=2$ provides the optimal trade-off between recall and efficiency in datasets dominated by small objects, and we therefore adopt it as the fixed parameter in the SPD module of TinyDef-DETR.
		
		\subsubsection{Design of the Subsequent Convolution}
		Following SPD, we adopt a carefully designed convolutional layer to refine the rearranged features. Specifically, we employ a non-strided convolution with a kernel size of $3 \times 3$. This configuration is chosen because the $3 \times 3$ kernel provides the minimal receptive field capable of covering the $2 \times 2$ local neighborhood that is redistributed during SPD. In contrast, a $1 \times 1$ kernel would only perform channel mixing without explicitly modeling cross-partition spatial relationships, which are essential for capturing subtle structural variations in small defects. Larger kernels such as $5 \times 5$ would increase computation without offering proportional benefits for objects of the scale encountered in transmission line inspection.
		
		We set the stride to $1$ in order to avoid any additional downsampling. Since the primary motivation of SPD is to preserve pixel-level detail while reducing resolution in a lossless manner, further spatial reduction at this stage would counteract the benefits of SPD and degrade the detection of small defects.
		
		In terms of channel configuration, the input dimensionality after SPD expands to four times the original number of channels, consistent with $\text{scale}=2$. This expansion ensures that all pixel-level information is preserved in the channel space. The output channel dimension is then set to match the base width of the subsequent stage in TinyDef-DETR, ensuring both computational efficiency and stable gradient propagation across the backbone.
		
		\subsection{Cross-Stage Dual-Domain Multi-Scale Attention Module}
		\label{sec:csddmsa}
		
		Transmission line defects in UAV imagery—such as cracks, rust spots, and insulator damages—are typically small, weakly salient, and embedded in cluttered backgrounds of vegetation, sky, or towers. Their ambiguous boundaries and subtle textural cues challenge conventional CNNs, whose local receptive fields hinder simultaneous modeling of global context and fine detail. Simply enlarging kernels inflates parameters and computation, often causing oversmoothing that erases critical structures\cite{finderWaveletConvolutionsLarge2025}. Likewise, attention mechanisms restricted to a single domain—spatial or channel—offer limited gains for tiny, low-contrast targets\cite{zhouUAVImageSmall2025}. Building on the joint frequency–spatial perspective outlined in Section~\ref{sec:Review of Joint Frequency–Spatial}, we introduce the \emph{Cross-Stage Dual-Domain Multi-Scale Attention Module} (CSDMAM), which unifies frequency-domain selectivity with spatial-domain localization in an end-to-end attention framework.
		
		The central idea is to embed learnable frequency gating directly within attention formation while retaining anisotropic, large-kernel spatial paths tailored to line-shaped defects. Frequency responses preserve edges, textures, and other high-frequency cues and suppress low-frequency clutter, whereas spatial aggregation localizes potential defects under complex context. Rather than treating frequency information as an auxiliary branch, CSDMAM integrates it adaptively so that frequency cues steer attention weights toward defect-relevant structures and away from distractors.
		
\begin{figure}[htb]
			\centering
			\includegraphics[width=0.8\textwidth]{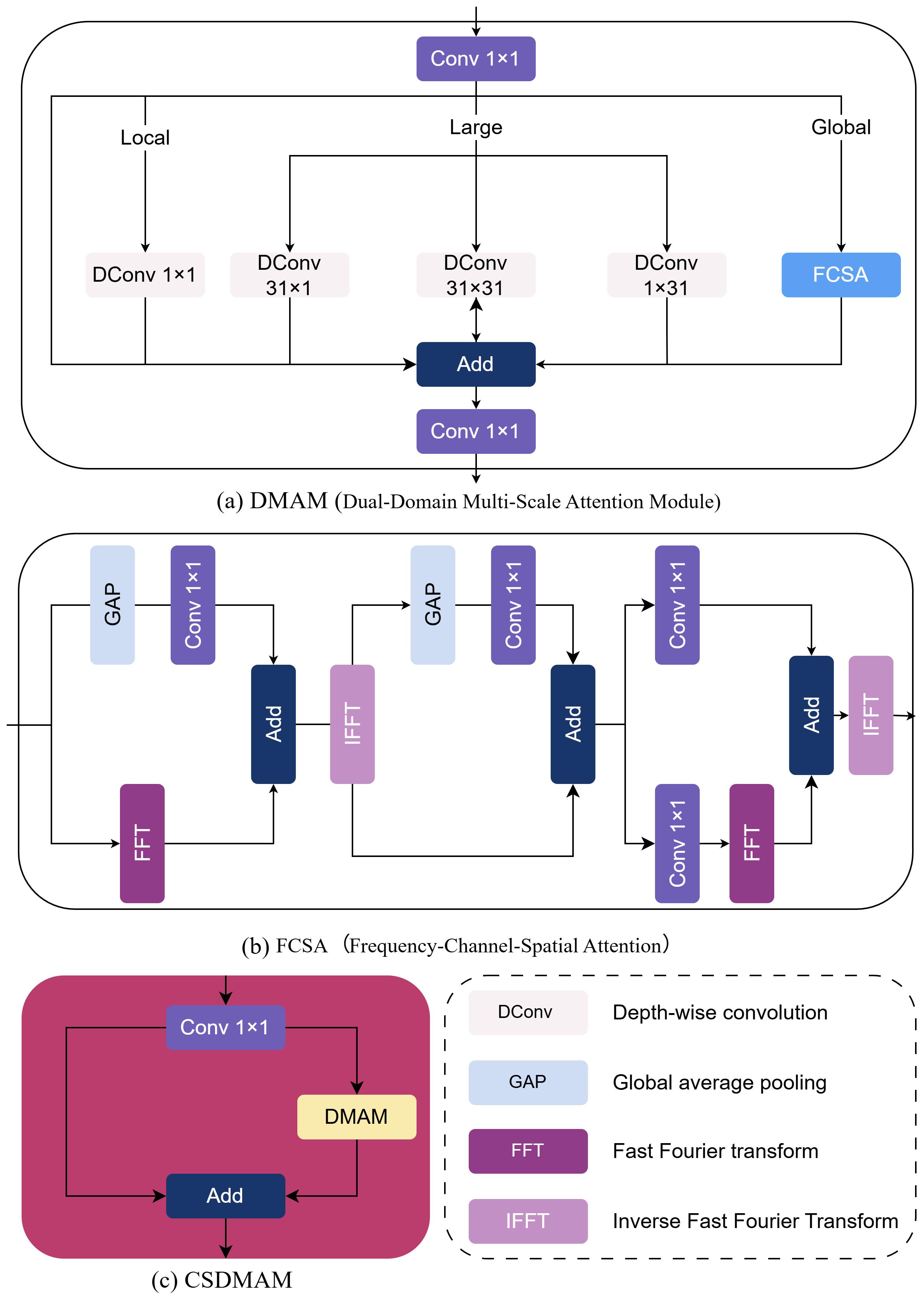}
			\caption{The architecture of the proposed Cross-Stage Dual-Domain Multi-Scale Attention Module (CSDMAM).}
			\label{fig:CSDMAM}
		\end{figure}

		Given an input feature map \(X \in \mathbb{R}^{B \times C \times H \times W}\), the module first applies a \(1\times 1\) convolution to adjust channel dimensionality. The transformed tensor then flows through several parallel, specialized branches that operate across complementary scales. One branch performs a learnable frequency decomposition whose gated responses accentuate sharp structural variations indicative of defects while attenuating smooth backgrounds; these gated maps participate directly in attention computation. In parallel, an anisotropic large-kernel spatial path—implemented with directional and dilated filters—approximates a wide receptive field without excessive parameters and mitigates oversmoothing, thereby enhancing sensitivity to elongated conductor- and insulator-like patterns. Additional scale-selective paths aggregate broader semantic context and preserve fine detail. The branch outputs are merged by residual summation and projected with a final \(1\times 1\) convolution, as illustrated in Figure~\ref{fig:CSDMAM}.
		
		Through this dual-domain, multi-scale design, CSDMAM preserves high-frequency, edge-dominated details, captures long-range semantics with modest overhead, and adaptively highlights subtle, defect-related responses. The resulting module addresses the principal failure modes of purely convolutional and single-domain attention approaches, delivering robust detection of small, low-saliency defects in complex aerial scenes.
		
		To effectively capture complementary information at different scales and semantic levels, CSDMAM is composed of four parallel branches: a Large-Receptive-Field Branch, a Local Detail Preservation Branch, a Global Branch based on the Frequency–Spatial–Channel Attention (FSCA) module, and a Residual Aggregation Branch.
		
		\textbf{Large-Receptive-Field Branch.}
		To enlarge the receptive field, we employ depthwise-separable convolutions with kernels \((k,k)\), \((1,k)\) and \((k,1)\), where \(k=31\). The square \(k\times k\) kernel aggregates global context, while the horizontal and vertical strip kernels capture elongated dependencies that are critical for the line-shaped structures typical of transmission infrastructure.
		The choice \(k=31\) was determined by systematic empirical evaluation rather than by heuristic selection (see Section~\ref{sec:exp-kernel-freq}). Kernels substantially smaller than \(31\) fail to capture the long-range, directionally elongated dependencies of line-like defects, whereas substantially larger kernels introduce excessive smoothing and disproportionate computational overhead without commensurate accuracy gains. Therefore, a \(31\times31\) depthwise-separable kernel provides the most favorable trade-off between global context aggregation and detail preservation for our task.
		We further restrict \(k\) to the family \(k=2^{n}-1\) (here \(31=2^{5}-1\)) for two practical reasons. First, an odd kernel size preserves a single central pixel, which simplifies symmetric, centered convolutional responses and padding. Second, when features are aggregated across hierarchical stages that expand spatial support by factors of two—via pooling, striding, or stage-wise downsampling—the geometric series \(\sum_{i=0}^{n-1}2^{i}=2^{n}-1\) naturally arises. Selecting \(k=2^{n}-1\) aligns the kernel's spatial support with an \(n\)-level binary-scale expansion, facilitating efficient cross-scale context aggregation and predictable padding behavior. The ablation in Section~\ref{sec:exp-kernel-freq} confirms that \(k=31\) is optimal in our application.
		
		\textbf{Local Detail Preservation Branch.}
		To prevent important fine-grained cues from being washed out by large-kernel operations, we introduce a lightweight local path using a depth-wise \(1\times1\) convolution, which preserves high-frequency details such as micro-cracks and corrosion spots while adding negligible computational burden.
		
		\textbf{Global Branch}
		To strengthen semantic discrimination and capture long-range contextual information, we design a global branch built upon the FSCA module that jointly exploits frequency-, spatial-, and channel-wise dependencies. Given the input \(X\), we first compute the frequency-domain representation
		\[
		X_f=\mathcal{F}(X),
		\]
		where \(\mathcal{F}\) denotes the discrete Fourier transform (DFT). Channel-wise weights are then obtained via global average pooling followed by a \(1\times1\) convolution, yielding \(A_c\in\mathbb{R}^{B\times C\times1}\). These weights modulate the frequency response,
		\[
		\tilde{X}_f=A_c\cdot X_f,
		\]
		which is transformed back to the spatial domain through the inverse DFT,
		\[
		X_{fs}=\mathcal{F}^{-1}(\tilde{X}_f).
		\]
		Finally, a spatial–channel reweighting map \(A_{sc}\) is generated via global pooling and a \(1\times1\) convolution to adaptively emphasize defect-relevant regions, producing the branch output
		\[
		X_{fsca}=A_{sc}\cdot X_{fs}.
		\]
		The effectiveness of FSCA and the integration of frequency-domain cues is validated in Section~\ref{sec:exp-kernel-freq}.
		
		\textbf{Residual Aggregation Branch.}
		To effectively integrate complementary information from all branches while preserving the original signals, we adopt a residual fusion mechanism that combines the outputs of the large-receptive-field branch, the local branch, and the FSCA branch with the input via element-wise summation:
		\[
		Y = X + DW_{1,k}(X_n) + DW_{k,1}(X_n) + DW_{k,k}(X_n) + DW_{1,1}(X_n) + X_{fsca},
		\]
		where \(DW\) denotes depth-wise convolution with the specified kernel size. This design enhances gradient flow, stabilizes optimization, and consolidates global context, fine details, and dual-domain attention responses into a unified representation for precise defect localization.
		
		To further improve computational efficiency and facilitate gradient propagation, we extend CSDMAM into a cross-stage structure following the Cross Stage Partial (CSP) strategy. The input is split into two parts: one processed by the multi-branch attention module and the other preserved as an identity mapping. The two parts are concatenated and fused by a \(1\times1\) convolution to produce the final output:
		\[
		Y_{cs}=\mathrm{Conv}\big([\mathrm{DMAM}(X_{\text{branch}}),\,X_{\text{identity}}]\big).
		\]
		By jointly leveraging large receptive fields, local detail preservation, FSCA-based dual-domain enhancement, and cross-stage fusion, CSDMAM achieves a favorable balance between global semantic understanding and precise defect localization, significantly improving the detection of small, visually ambiguous defects in UAV inspection imagery.
		
		\subsection{Focaler–Wise–SIoU Loss}
		\label{sec:loss}
		
		In transmission-line defect detection, targets are extremely small and exhibit subtle spatial variations, so precise bounding-box regression is pivotal for reliable recognition. IoU-based losses are attractive for their geometric interpretability and optimization efficiency, yet the canonical IoU measures only overlap and provides no informative gradients when the boxes are disjoint, which stalls learning on difficult examples. Subsequent variants—GIoU \cite{rezatofighiGeneralizedIntersectionUnion2019a}, DIoU \cite{zhengDistanceIoULossFaster2020}, and CIoU \cite{zhengEnhancingGeometricFactors2022}—mitigate this to some extent by adding enclosure, center-distance, and aspect-ratio terms. Nevertheless, they still struggle to capture spatial alignment and shape consistency in a unified manner. More advanced designs such as SIoU \cite{gevorgyanSIoULossMore2022} integrate angular, distance, and shape components, but typically weight all samples uniformly; gradients then become dominated by easy cases while hard, tiny, low-SNR objects receive insufficient emphasis. Focaler–IoU \cite{zhangFocalerIoUMoreFocused2024b} further reshapes the overlap signal by linearly remapping IoU within a fixed interval to favor certain ranges, but its emphasis is uniform and monotonic and it lacks angle-aware distance/shape coupling. Wise-IoU \cite{tongWiseIoUBoundingBox2023a} introduces a dynamic, non-monotonic focusing mechanism that suppresses low-quality examples, yet its default formulation is ill-suited to our setting: it deemphasizes geometric shape and aspect consistency that are crucial for sub-pixel localization of tiny defects, it normalizes outlier degree at the batch level—which can miscalibrate under low-SNR, highly imbalanced anchors and inadvertently down-weight informative hard-but-valid samples—and it omits explicit angle-aware distance coupling, yielding weaker guidance for axis-aligned center drift near overlap. Compounding these issues, conventional IoU-based objectives offer limited discrimination in the high-IoU regime where fine-grained localization matters most, leading to slow convergence and reduced accuracy.
		
		To address these shortcomings, we propose Focaler–Wise–SIoU, a unified regression objective that couples normalized IoU scaling, comprehensive geometric penalties, and difficulty-aware modulation. The goal is to sustain strong, stable gradients across the entire overlap spectrum while explicitly promoting spatial alignment and shape fidelity for tiny objects. Concretely, we first introduce a Focaler-IoU overlap surrogate that clips and linearly rescales IoU to sharpen gradients near perfect alignment, while preserving the raw IoU for principled reweighting. We then design SIoU-style geometric penalties that combine an angle-aware, axis-normalized center-distance term with a smooth, saturating shape discrepancy, encouraging accurate alignment and aspect consistency under tiny-object noise. Finally, we incorporate a non-monotonic focal modulation (Wise) driven by an EMA-normalized IoU gap, which highlights informative, moderately hard samples and down-weights both trivial cases and extreme outliers. Raw IoU is used exclusively for difficulty estimation to retain geometric interpretability, and the overall objective remains closed-form, differentiable, and implementation-friendly. These components jointly accelerate convergence and substantially improve localization accuracy on small, low-SNR defects. 
		
		The following subsections introduce notation, define each component with its rationale, and present the final loss composition.
		
		\subsubsection{Preliminaries and Notation}
		\label{sec:preliminaries}
		
		Let the predicted and target bounding boxes be
		\(
		B_p=(x_p,y_p,w_p,h_p)\) and \(B_t=(x_t,y_t,w_t,h_t)\),
		parameterized by center coordinates \((x,\!y)\) and width/height \((w,\!h)\).
		We write the top-left / bottom-right corners as
		\[
		\begin{aligned}
			x_p^{(1)}&=x_p-\tfrac{w_p}{2},\quad & y_p^{(1)}&=y_p-\tfrac{h_p}{2},\quad
			& x_p^{(2)}&=x_p+\tfrac{w_p}{2},\quad & y_p^{(2)}&=y_p+\tfrac{h_p}{2},\\
			x_t^{(1)}&=x_t-\tfrac{w_t}{2},\quad & y_t^{(1)}&=y_t-\tfrac{h_t}{2},\quad
			& x_t^{(2)}&=x_t+\tfrac{w_t}{2},\quad & y_t^{(2)}&=y_t+\tfrac{h_t}{2}.
		\end{aligned}
		\]
		The intersection size is
		\begin{align}
			w_{\mathrm{int}} &= \max\!\bigl(0,\,\min(x_p^{(2)},x_t^{(2)})-\max(x_p^{(1)},x_t^{(1)})\bigr),\\
			h_{\mathrm{int}} &= \max\!\bigl(0,\,\min(y_p^{(2)},y_t^{(2)})-\max(y_p^{(1)},y_t^{(1)})\bigr).
		\end{align}
		with area \(S_{\mathrm{int}}=w_{\mathrm{int}}h_{\mathrm{int}}\). The union area
		\(S_{\mathrm{union}}=w_p h_p+w_t h_t-S_{\mathrm{int}}\), and the standard IoU is
		\begin{equation}
			\mathrm{IoU}=\frac{S_{\mathrm{int}}}{S_{\mathrm{union}}}\in[0,1].
			\label{eq:std_iou}
		\end{equation}
		We also denote the center displacement by
		\(
		\Delta=(d_x,d_y)=(x_p-x_t,\,y_p-y_t)
		\),
		and the side lengths of the minimum enclosing rectangle of \(B_p\) and \(B_t\) by
		\begin{equation}
			(w_{\mathrm{box}},h_{\mathrm{box}})=
			\bigl(\max(x_p^{(2)},x_t^{(2)})-\min(x_p^{(1)},x_t^{(1)}),\;
			\max(y_p^{(2)},y_t^{(2)})-\min(y_p^{(1)},y_t^{(1)})\bigr).
			\label{eq:enclosing}
		\end{equation}
		\emph{Numerical stability.} We use small constants where needed: \(10^{-4}\) in Eq.~\eqref{eq:phi} and \(\varepsilon=10^{-6}\) in Sec.~\ref{sec:wise} to avoid singularities; the EMA momentum \(m=10^{-2}\) balances variance and adaptability across training phases.
		
		\subsubsection{Focaler-IoU Overlap Term}
		\label{sec:focaler}
		
		In transmission-line defect detection most objects are tiny; training benefits from non-saturating gradients even at high overlaps. To strengthen gradient sensitivity near perfect alignment while keeping the geometric meaning of Eq.~\eqref{eq:std_iou} for reweighting, we use a \emph{clipped and linearly rescaled proxy} \(\hat{\mathrm{IoU}}\) \emph{only} inside the overlap penalty:
		\begin{equation}
			\hat{\mathrm{IoU}}=\mathrm{clip}\!\left(\frac{\mathrm{IoU}-d}{u-d},\,0,\,1\right),\qquad (d,u)=(0,\,0.95),
			\label{eq:focaler_iou}
		\end{equation}
		and define
		\begin{equation}
			L_{\mathrm{IoU}}=1-\hat{\mathrm{IoU}}.
			\label{eq:lioU}
		\end{equation}
		\emph{High-IoU sensitivity.} Using \(\hat{\mathrm{IoU}}\) solely in \(L_{\mathrm{IoU}}\) preserves gradient magnitude as \(\mathrm{IoU}\!\to\!1\) while the \emph{raw} IoU (Eq.~\eqref{eq:std_iou}) is retained for difficulty modulation, maintaining geometric interpretability.
		
		\subsubsection{Angle-Weighted Distance Penalty}
		\label{sec:angle_distance}
		
		In transmission line defect detection tasks, objects such as insulator cracks, missing bolts, or bird nests often appear at arbitrary orientations and with substantial variation in scale and spatial distribution. These defects are typically small, elongated, and densely distributed along complex backgrounds such as towers or cables, especially in aerial inspection imagery. In such conditions, even minor center displacements between predicted and ground-truth bounding boxes can result in significant localization errors, particularly when the offset directions are not well aligned. Conventional distance-based penalties treat all directions uniformly—whether horizontal, vertical, or diagonal—thereby neglecting the geometric structure inherent in these domain-specific misalignments. 
		
		This limitation motivates us to develop an angle-weighted distance penalty that dynamically adjusts the localization penalty according to the angular alignment of the offset vector. By doing so, the model is encouraged to favor axis-aligned center corrections over diagonal ones, improving directional sensitivity and enhancing localization accuracy under the challenging settings encountered in power line defect detection.
		
		We define the offset between the predicted and ground-truth box centers as \(\Delta = (d_x, d_y)\), and let \(\phi \in [0, \pi/4]\) denote the acute angle between \(\Delta\) and its nearest coordinate axis. This angle reflects the degree of directional misalignment: \(\phi = 0\) indicates perfect alignment with one axis, while \(\phi = \pi/4\) corresponds to a fully diagonal offset. It is computed as:
		\begin{equation}
			\phi = \arcsin\left(\frac{\min(|d_x|, |d_y|)}{\sqrt{d_x^2 + d_y^2} + 10^{-4}}\right),
			\label{eq:phi}
		\end{equation}
		where a small constant in the denominator ensures numerical stability near zero displacement.
		
		In the original SIoU formulation, the angle cost is defined as:
		\[
		\Lambda = 1 - 2\sin^2\left(\phi - \tfrac{\pi}{4}\right) = \sin(2\phi),
		\]
		which increases as the offset becomes more axis-aligned. When the displacement vector approaches either the horizontal or vertical axis, the angle \(\phi\) becomes small, resulting in a higher \(\Lambda\), indicating better alignment between the predicted and target boxes.
		
		To incorporate this angular alignment into the penalty formulation, we define a scaling factor that modulates the influence of the distance terms:
		\begin{equation}
			\mathrm{Angle} = \sin(2\phi) - 2 = - (2 - \Lambda),
			\label{eq:Angle}
		\end{equation}
		which ensures that \(\mathrm{Angle} \le 0\). This negative-valued quantity is then used in an exponential decay to penalize displacement, such that larger directional misalignment results in steeper penalties. Specifically, when the offset direction is diagonal (i.e., small \(\Lambda\)), the penalty grows more aggressively; when it is axis-aligned (i.e., large \(\Lambda\)), the penalty grows more slowly. This behavior makes the model more sensitive to the type of misalignment rather than only its magnitude.
		
		Based on this principle, we formulate the distance penalty as:
		\begin{equation}
			\mathrm{Dist} = 2 
			- \exp\left(\mathrm{Angle} \cdot \left(\frac{d_x}{w_{\mathrm{box}}}\right)^2\right) 
			- \exp\left(\mathrm{Angle} \cdot \left(\frac{d_y}{h_{\mathrm{box}}}\right)^2\right),
			\label{eq:dist}
		\end{equation}
		where \(w_{\mathrm{box}}\) and \(h_{\mathrm{box}}\) represent the width and height of the smallest enclosing box that covers both the predicted and ground-truth boxes. Each exponential term decreases as the normalized offset increases due to the negative scaling factor, ensuring that \(\mathrm{Dist}\) is monotonic in misalignment. Moreover, axis-wise normalization promotes scale invariance, which is crucial for handling object size variations in diverse datasets.
		
		In summary, this angle-weighted distance penalty effectively captures both the magnitude and direction of misalignment. It provides stronger supervision in challenging cases such as small, rotated, or overlapping objects, as commonly observed in the DIOR and VisDrone datasets. Empirically, we find that incorporating this term improves detection accuracy by enhancing the localization sensitivity to angular deviation.
		
		\subsubsection{Shape Consistency Penalty}
		\label{sec:shape}
		
		Tiny defects often suffer from annotation noise; a saturating yet smooth transform mitigates the dominance of extremely large aspect errors while keeping sensitivity to moderate mismatches. We adopt
		\begin{equation}
			\mathrm{Shape}=\Bigl(1-\exp\!\bigl(-\tfrac{|w_p-w_t|}{\max(w_p,w_t)}\bigr)\Bigr)^{\theta}
			+\Bigl(1-\exp\!\bigl(-\tfrac{|h_p-h_t|}{\max(h_p,h_t)}\bigr)\Bigr)^{\theta},\qquad \theta=4.
			\label{eq:shape}
		\end{equation}
		For small relative deviations, \(1-\exp(-x)\approx x\) is near-linear; for large deviations it saturates towards~1, and the exponent \(\theta\) further curbs outliers while preserving gradients for mid-range shape errors. This saturating design improves \emph{robustness to annotation noise} on tiny objects without letting large, noisy deviations dominate the loss.
		
		Combining the overlap penalty (Eq.~\eqref{eq:lioU}) with the distance and shape terms (Eqs.~\eqref{eq:dist}--\eqref{eq:shape}), we obtain the SIoU-style geometric core:
		\begin{equation}
			L_{\mathrm{SIoU}} \;=\; L_{\mathrm{IoU}} \;+\; \tfrac{1}{2}\bigl(\mathrm{Dist}+\mathrm{Shape}\bigr).
			\label{eq:lsiou_core}
		\end{equation}
		The factor \(1/2\) balances alignment/shape against overlap so that none dominates the gradient field in typical training regimes. Note again that only \(L_{\mathrm{IoU}}\) uses the rescaled \(\hat{\mathrm{IoU}}\); all reweighting below relies on the \emph{raw} IoU.
		
		\subsubsection{Wise Difficulty Modulation via EMA Normalization}
		\label{sec:wise}
		
		Precise regression of small defects in UAV imagery is hindered by an imbalance in sample difficulty: abundant trivial cases dominate gradients, while naïve hard mining overfits noisy outliers and mislabels. We therefore propose Wise Difficulty Modulation (WDM)—a normalization-driven reweighting scheme that scales each sample’s gradient by an IoU-based difficulty estimate. WDM stabilizes training via an EMA-normalized difficulty distribution that adapts as accuracy improves, and uses a non-monotonic modulation to emphasize moderately hard, informative samples while down-weighting both trivial and extreme cases. This yields smoother optimization and consistently better localization, especially for small or ambiguous defects common in UAV datasets.
		
		We quantify sample difficulty using the normalized IoU gap:
		\begin{equation}
			\beta = \frac{1 - \mathrm{IoU}}{\mathbb{E}[\,1 - \mathrm{IoU}\,] + \varepsilon},
			\qquad
			\mathbb{E}[\,1 - \mathrm{IoU}\,] \leftarrow (1 - m)\,\mathbb{E}[\,1 - \mathrm{IoU}\,] + m \cdot (1 - \mathrm{IoU})_{\text{batch}},
			\label{eq:beta}
		\end{equation}
		where the expectation is tracked by an exponential moving average (EMA). We use \(m=10^{-2}\), which provides a stable yet responsive adaptation to phase shifts during training; in practice, a tiny constant \(\varepsilon=10^{-6}\) is added to the denominator for numerical stability. Crucially, Eq.~\eqref{eq:beta} operates on the \emph{raw} IoU from Eq.~\eqref{eq:std_iou} to preserve geometric interpretability. This normalization eliminates the influence of absolute IoU scale drift, allowing difficulty to be compared consistently across epochs and batch conditions.
		
		To emphasize moderately hard examples while avoiding overfitting to extreme outliers, we employ the following modulation:
		\begin{equation}
			\gamma(\beta)=\frac{\beta}{\delta\,\alpha^{(\beta-\delta)}},\qquad \alpha=1.7,\ \delta=2.7.
			\label{eq:gamma}
		\end{equation}
		
		This non-monotonic weighting concentrates higher weights on samples with moderate normalized difficulty while gently down-weighting trivially easy cases and preventing domination by extreme outliers as $\beta$ increases. Empirically, such weighting increases the proportion of informative samples contributing to effective gradient updates, accelerating convergence and improving the recall of subtle, low-SNR targets without introducing instability.
		
%
		
		Overall, Wise Difficulty Modulation via EMA Normalization offers a self-adaptive, statistically grounded mechanism to stabilize training under severe difficulty imbalance. In the context of transmission line defect datasets, it effectively improves localization precision and reduces missed detections of small and ambiguous defects by ensuring that learning focuses on the most informative samples throughout the optimization process.
		
		\subsubsection{Final Objective and Batch-wise Computation}
		\label{sec:final}
		
		Multiplying the geometric core (Eq.~\eqref{eq:lsiou_core}) by the difficulty weight (Eq.~\eqref{eq:gamma}) yields the \emph{Focaler--Wise--SIoU} objective. 
		Recalling that
		\[
		L_{\mathrm{IoU}} = 1 - \hat{\mathrm{IoU}}, 
		\qquad 
		L_{\mathrm{SIoU}} = L_{\mathrm{IoU}} + \tfrac{1}{2}\bigl(\mathrm{Dist}+\mathrm{Shape}\bigr),
		\]
		the final loss can be compactly written as
		\begin{equation}
			\boxed{
				L \;=\; L_{\mathrm{SIoU}}\,\gamma(\beta),
				\quad
				\beta=\frac{1-\mathrm{IoU}}{\mathbb{E}[\,1-\mathrm{IoU}\,]+\varepsilon}
			}
			\label{eq:final_loss}
		\end{equation}
		where $\beta$ is defined from the raw IoU as in Eq.~\eqref{eq:beta}, and $\gamma(\beta)$ is given in Eq.~\eqref{eq:gamma}. 
		The EMA-based normalization acts only through $\beta$, and no additional normalization is introduced in the final objective.
		\section{Results}
		\subsection{Datasets}
		\label{sec:dataset}
		
		In this paper, we use the CSG-ADCD dataset. The CSG-ADCD (China Southern Grid Aerial Defective Component Dataset) is a large-scale, high-resolution UAV-based image dataset specifically developed for the detection of defects in power transmission line components, including insulators, tie wires, and poles. Constructed under authentic inspection scenarios by China Southern Power Grid, the dataset comprises 10,000 aerial images. Each image was manually annotated by a single domain expert to ensure both consistency and high-quality labeling.
		
		In total, the dataset contains 73,448 annotated instances covering nine component states or defect types: Normal Insulator (zcjyz), Polluted Insulator (jyzwh), Damaged Pole (dgss), Missing Tie Wire (zxqs), Loose Tie Wire (zxst), Insulator Flashover (jyzsl), Bird Nest (nw), Broken Insulator (jyzps), and Shattered Post Insulator (zyzsl). 
		
		Table \ref{tab:label_category} provides a correspondence between the labels and category names for each defect type in the dataset, offering a clear overview of the classification scheme used.
		
		On average, each object occupies a pixel area of 422.12, with the dataset exhibiting a highly imbalanced distribution in terms of object sizes: 94.51\% small objects (69,413 instances), 5.48\% medium-sized (4,022 instances), and only 0.02\% large (13 instances). This reflects the intrinsic characteristics of UAV-based transmission line inspection objects, which are often small, localized, and partially occluded.
		
		For training consistency, all images were preprocessed by resizing the longer side to 640 pixels while preserving aspect ratio, followed by padding to a fixed resolution of 640 × 640 pixels using gray borders. The dataset is split into 8,000 training, 1,000 validation, and 1,000 test images.
		
		\begin{table}[htb]
			\centering
			\caption{Correspondence between labels and category names}
			\vspace{-12pt}
			\label{tab:label_category}
			\begin{tabularx}{\textwidth}{YYY}  
				\toprule
				No. & Label & Name \\
				\midrule
				1 & zcjyz & Normal Insulator \\
				2 & jyzwh & Polluted Insulator \\
				3 & dgss  & Damaged Pole \\
				4 & zxqs  & Missing Tie Wire \\
				5 & zxst  & Loose Tie Wire \\
				6 & jyzsl & Insulator Flashover \\
				7 & nw    & Bird Nest \\
				8 & jyzps & Broken Insulator \\
				9 & zyzsl & Shattered Post Insulator \\
				\bottomrule
			\end{tabularx}
		\end{table}
		
		Statistical analysis further highlights two key features of the dataset. First, the scatter distribution of object width and height reveals a strong concentration of samples in the lower-left region, confirming the overwhelming dominance of small-scale objects. Second, the number of objects per image follows a relatively stable distribution, averaging 7.34 instances per image, with 2,285 images containing between 7 and 8 objects. This regularity is particularly beneficial for developing models optimized for dense and small-object detection.
		
		In summary, the CSG-ADCD dataset demonstrates strong representativeness by simultaneously addressing two critical aspects. On the one hand, it captures a comprehensive range of component defects and status indicators that are directly relevant to transmission safety, while also reflecting the complexity of real-world inspection conditions such as occlusion, cluttered backgrounds, and varying illumination. On the other hand, its pronounced dominance of small-scale objects—over 94\% of all annotated instances—together with the presence of subtle and fine-grained anomalies, makes it a highly challenging and domain-specific benchmark. In this regard, CSG-ADCD not only advances research in defect detection for power systems but also provides a valuable testbed for generic small-object detection in aerial remote sensing, offering greater domain relevance than existing datasets such as VisDrone\citep{duVisDroneDET2019VisionMeets2019}.
		
		Overall, CSG-ADCD constitutes a robust and high-fidelity benchmark, offering significant value for advancing research in both domain-specific defect detection and general small-object detection tasks.
		
\begin{figure}[htb]
			\centering
			\includegraphics[width=0.8\textwidth]{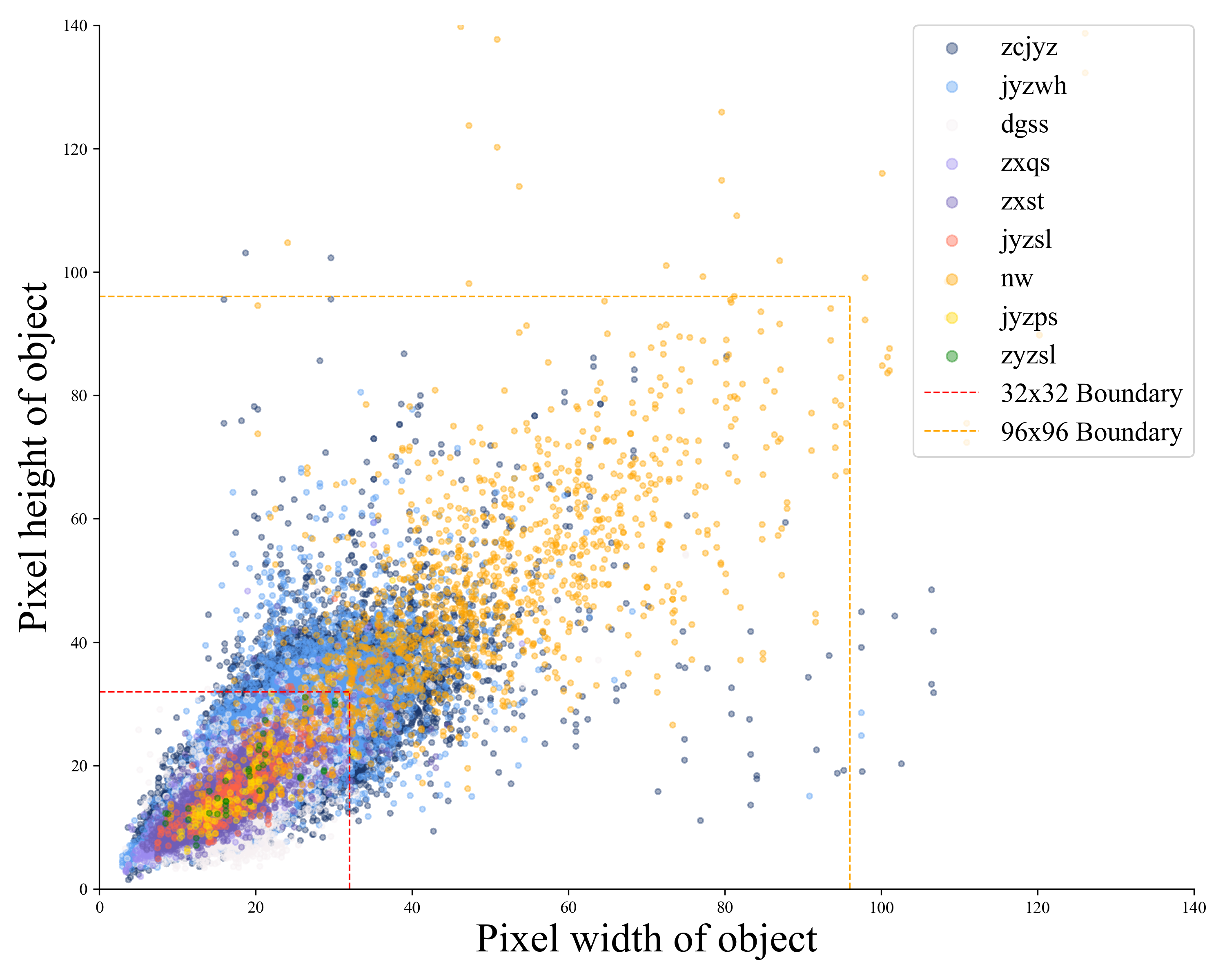}
			\caption{Distribution of object sizes in CSG-ADCD}
			\label{fig:Distribution}
		\end{figure}

		As shown in Fig.\ref{fig:Distribution}, the vast majority of objects are smaller than 32×32 pixels, placing them within the small-object category. The scatter plot depicts the distribution of annotated instances by pixel width and height after preprocessing, with different categories distinguished by distinct colors in the legend. The strong concentration of points in the lower-left corner highlights the predominance of small and compact objects, consistent with the characteristics of transmission-line inspection imagery. Dashed reference lines at 32×32 and 96×96 pixels mark the boundaries between small, medium, and large regimes, clearly revealing the relative scarcity of medium- and large-sized objects. 
		
		Overall, the visualization emphasizes the dominance of small instances and underscores the dense, fine-grained nature of aerially captured transmission-line defects.
		
		\subsection{Model Training and Evaluation Metrics}
		
		The proposed model was implemented in the PyTorch framework and built on top of the Ultralytics RT-DETR backbone. Experiments were conducted on a workstation equipped with an NVIDIA GeForce RTX 3090 (24\,GB) running Ubuntu 20.04, Python 3.10.14, PyTorch 2.3.1, and CUDA 12.1. Training proceeded for 100 epochs with a batch size of 4 and an input resolution of $640\times640$. Optimization employed AdamW with an initial learning rate of $1.0\times10^{-4}$, a final learning-rate factor $lrf=1.0$, momentum $0.9$, and weight decay $1.0\times10^{-4}$. We used a 2{,}000-iteration warm-up during which the optimizer momentum was set to $0.8$ and the bias learning rate was scaled by a factor of $0.1$.
		
		Evaluation is conducted on the \textit{test} split following the COCO evaluation protocol, where we report mAP$_{50}$, the mean Average Precision at an IoU threshold of 0.50, and mAP$_{50:95}$, the mean Average Precision averaged across IoU thresholds from 0.50 to 0.95 in increments of 0.05. During inference, class-agnostic non-maximum suppression (NMS) with an IoU threshold of 0.7 is applied, retaining at most 300 detections per image. Note that this NMS threshold is independent of the IoU thresholds used for AP and mAP computation.
		
		The COCO protocol additionally provides scale-sensitive metrics that illuminate detector performance across object sizes: $AP_s$ measures average precision for small objects (area $<32^2$ pixels) and $AP_m$ measures average precision for medium objects (area between $32^2$ and $96^2$ pixels). Because most defects in transmission-line inspection datasets are small, $AP_s$ is especially informative of the model’s ability to detect fine-grained, low–pixel-count anomalies.
		
		We evaluate inference efficiency and model complexity using standard measures: floating-point operations (FLOPs) quantify computational complexity and model size is reported as the number of parameters (Params, in millions). Detection performance is analyzed using precision ($P$), recall ($R$), average precision (AP), and mean average precision (mAP). These metrics are defined as
		\[
		P=\frac{TP}{TP+FP}, \qquad
		R=\frac{TP}{TP+FN},
		\]
		where $TP$, $FP$, and $FN$ denote true positives, false positives, and false negatives, respectively. The average precision for a single category is computed as the area under the precision–recall curve,
		\[
		AP=\int_{0}^{1} P(R)\,\mathrm{d}R,
		\]
		and the mean average precision across $k$ categories is given by
		\[
		\mathrm{mAP}=\frac{1}{k}\sum_{i=1}^k AP_i.
		\]
		
		To provide a more fine-grained assessment relevant to transmission-line inspection, we additionally report scale-sensitive metrics ($AP_s$, $AP_m$) and category-specific results for the two most prevalent defect types: Polluted Insulator (jyzwh) and Damaged Pole (dgss). For these categories we present both recall and $\mathrm{mAP}_{50}$ to highlight the model’s ability to both locate and precisely localize commonly occurring, safety-critical defects. Together, the described computational and detection metrics provide a balanced appraisal of accuracy, completeness, and operational efficiency for real-time UAV-based transmission-line inspection.
		
		Finally, consistent with the above accuracy–efficiency analyses and evaluation protocol, we emphasize the intended deployment mode: post-mission processing of UAV imagery on a server (rather than on-board inference). Using the same $640\times640$ input as in our comparisons, TinyDef-DETR sustains \textbf{152.19~FPS} on a single NVIDIA RTX~3090 (24\,GB)—that is, \textbf{152.19 images/s} or \textbf{6.57\,ms} per image ($1000/152.19\!\approx\!6.57$\,ms). This measured throughput is well aligned with the accuracy–complexity trade-offs reported above and is operationally meaningful: a typical batch of 1{,}000 UAV images can be processed in approximately \textbf{6.57\,s} of pure inference time, readily meeting the requirements of server-side, after-mission defect screening for transmission-line inspection.
		
		\subsection{Effects of Kernel Size and Frequency-Domain Features on Detection Performance}
		\label{sec:exp-kernel-freq}
		
		We quantitatively assessed the impact of the convolutional kernel size in the Large-Receptive-Field branch of CSDMAM (Section~\ref{sec:csddmsa}) by sweeping \(k\in\{3,7,15,31,63,127\}\). Table~\ref{tab:kernel-size} summarizes model complexity (Params, GFLOPs) and detection metrics (Precision, Recall, mAP\(_{50}\), mAP\(_{50:95}\), AP\(_s\), AP\(_m\)), including the newly added configuration \(k=127\).
		
		A consistent trend emerges. Very small kernels (\(k=3\)) under-capture long-range structure: despite minimal complexity they yield markedly lower Recall and mAP\(_{50}\) on elongated, line-like defects (Recall \(=0.206\), mAP\(_{50}=0.184\)). Moving to moderate sizes (\(k=7,15\)) modestly improves contextual aggregation and occasionally Precision, but the gains in Recall and AP remain limited, indicating insufficient coverage of the anisotropic spatial extents typical of transmission-line defects.
		
		Expanding to \(k=31\) provides the best overall balance. This setting delivers the highest Recall (\(0.263\)) and the strongest mAP\(_{50}\) (\(0.275\)) among all candidates, with clear improvements for small and medium objects (AP\(_s=0.106\), AP\(_m=0.110\)) at only a marginal computational increase (GFLOPs \(\approx 65.3\)). We attribute these gains to a receptive field that is large enough to aggregate global line-structure context while still preserving fine detail, aided by the depthwise/separable parameterization and the frequency-guided FSCA path that stabilizes high-frequency cues without amplifying noise.
		
		Further enlarging the kernel beyond this point does not help. At \(k=63\), Precision inches up (0.589) but Recall and small/medium APs drop; at \(k=127\), Precision peaks (0.627) yet Recall falls to \(0.192\) and overall AP degrades (mAP\(_{50}=0.209\), mAP\(_{50:95}=0.0933\)). We observe two coupled effects: (i) \emph{oversmoothing} of fine structures caused by excessively wide spatial averaging, which suppresses weak, small targets emphasized by FSCA; and (ii) \emph{optimization and efficiency penalties}—GFLOPs rise from 65.3 (at \(k=31\)) to 84.1 (at \(k=127\)), with diminishing returns and a measurable recall deficit on subtle defects.
		
		Taken together, the ablation supports selecting \(k=31\) for the Large-Receptive-Field branch of CSDMAM. This configuration yields the most favorable trade-off among global context capture, preservation of fine-grained defect morphology, and computational efficiency in UAV-based transmission-line inspection. Frequency-domain enhancement (FSCA) remains beneficial across settings, but its effect is maximized when paired with a \emph{moderately large}, not extreme, spatial kernel—precisely the regime at \(k=31\).
		
		\vspace{-8pt}
		
		\setcounter{table}{1}
		\begin{table}[H]
			\centering
			\caption{Performance comparison for different kernel sizes in the Large-Receptive-Field branch.}
			\vspace{-16pt}
			\label{tab:kernel-size}
			\begin{tabularx}{\textwidth}{*{9}{>{\centering\arraybackslash}X}}
				\toprule
				Kernel Size & Params (M) & GFLOPs & Precision & Recall & mAP$_{50}$ & mAP$_{50:95}$ & AP$_s$ & AP$_m$ \\
				\midrule
				3   & \textbf{20.4} & \textbf{64.0} & 0.406 & 0.206 & 0.184 & 0.0866 & 0.082 & 0.056 \\
				7   & \textbf{20.4} & 64.1 & 0.603 & 0.210 & 0.224 & 0.1032 & 0.089 & 0.056 \\
				15  & \textbf{20.4} & 64.3 & 0.496 & 0.208 & 0.213 & 0.0988 & 0.089 & 0.091 \\
				31  & 20.5 & 65.3 & 0.534 & \textbf{0.263} & \textbf{0.275} & \textbf{0.1187} & \textbf{0.106} & \textbf{0.110} \\
				63  & 20.8 & 69.0 & 0.589 & 0.220 & 0.244 & 0.1108 & 0.098 & 0.062 \\
				127 & 22.0 & 84.1 & \textbf{0.627} & 0.192 & 0.209 & 0.0933 & 0.083 & 0.058 \\
				\bottomrule
			\end{tabularx}
		\end{table}
		
		\vspace{-24pt}
		
		To comprehensively assess the impact of the FSCA branch and the integration of frequency-domain information, we evaluate three variants of our framework: TinyDef-DETR (w/o), where the FSCA branch and all frequency-domain operations are removed; TinyDef-DETR (Id), where the FSCA branch is retained but replaced with a pure identity mapping that forwards features without any modulation or frequency-domain interaction; and TinyDef-DETR (Full), which incorporates the complete FSCA-based dual-domain enhancement. The TinyDef-DETR (Id) variant is deliberately constructed as a neutral-control configuration. It is analogous to a biomedical trial investigating the effect of a soft drink on human health, in which one group receives the soft drink while another receives an equivalent volume of plain water. In our setting, TinyDef-DETR (Id) maintains the architectural footprint and computational cost of the FSCA branch while eliminating its adaptive behavior, thereby enabling a clean isolation of the genuine contribution of FSCA-based modulation from the mere presence of an additional branch.
		
		As reported in Table~\ref{tab:overall_performance}, TinyDef-DETR (Full) achieves the most favorable trade-off between detection accuracy and computational complexity. Compared with TinyDef-DETR (w/o), it consistently improves both Precision and Recall, with mAP$_{50}$ rising from 0.206 to 0.275 and mAP$_{50:95}$ increasing from 0.0926 to 0.1187, while adding only negligible overhead in parameters and GFLOPs. Although TinyDef-DETR (Id) attains the highest Precision (0.597), its mAP$_{50}$ (0.232) and mAP$_{50:95}$ (0.1015) remain clearly below those of TinyDef-DETR (Full). This discrepancy indicates that simply introducing an identity shortcut under the same computational budget cannot explain the observed performance gains; instead, they stem from the adaptive dual-domain feature modulation enabled by the FSCA mechanism.
	
		The per-class and scale-aware results in Table~\ref{tab:per_class_metrics} further reinforce this conclusion. For challenging categories such as \texttt{jyzwh} and \texttt{dgss}, TinyDef-DETR (Full) achieves higher Recall and mAP$_{50}$ than both TinyDef-DETR (w/o) and TinyDef-DETR (Id), indicating enhanced robustness to subtle, low-contrast, and visually ambiguous defects. In addition, TinyDef-DETR (Full) obtains the highest AP$_s$ and AP$_m$, confirming that the proposed dual-domain attention mechanism is effective in preserving fine-grained structural cues for small defects while simultaneously capturing richer contextual information for medium-scale targets. By contrast, TinyDef-DETR (Id) does not exhibit comparable improvements, which substantiates that the performance gains are attributed to the FSCA-based frequency-domain enhancement rather than to increased model capacity or superficial architectural modifications.
		
		\setcounter{table}{2}
		\begin{table}[htb]
			\centering
			\caption{Comparison of TinyDef-DETR variants with and without FSCA and frequency-domain enhancement. TinyDef-DETR (w/o): w/o FSCA; TinyDef-DETR (Id): Identity mapping.}
			\label{tab:overall_performance}
			\begin{tabular*}{\textwidth}{@{\extracolsep{\fill}} c c c c c c c @{}}
				\toprule
				Model & Params (M) & GFLOPs & Precision & Recall & mAP$_{50}$ & mAP$_{50:95}$ \\
				\midrule
				TinyDef-DETR (w/o)   & \textbf{20.4} & \textbf{64.3} & 0.465 & 0.208 & 0.206 & 0.0926 \\
				TinyDef-DETR (Full)~ & 20.5 & 65.3 & 0.534 & \textbf{0.263} & \textbf{0.275} & \textbf{0.1187} \\
				TinyDef-DETR (Id)~~~~& 20.5 & 65.3 & \textbf{0.597} & 0.222 & 0.232 & 0.1015 \\
				\bottomrule
			\end{tabular*}
		\end{table}
		
		\setcounter{table}{3}
		\begin{table}[htb]
			\centering
			\caption{Per-class Recall and mAP$_{50}$ for key defect categories and scale-wise AP. TinyDef-DETR (w/o): w/o FSCA; TinyDef-DETR (Id): Identity mapping.}
			\label{tab:per_class_metrics}
			\begin{tabular}{c c c c c c}
				\toprule
				Model & Recall (\texttt{jyzwh}) & mAP$_{50}$ (\texttt{jyzwh}) & mAP$_{50}$ (\texttt{dgss}) & AP$_s$ & AP$_m$ \\
				\midrule
				TinyDef-DETR (w/o)   & 0.369 & 0.337 & 0.389 & 0.081 & 0.060 \\
				TinyDef-DETR (Full)~ & \textbf{0.397} & \textbf{0.391} & 0.419 & \textbf{0.106} & \textbf{0.110} \\
				TinyDef-DETR (Id)~~~~& 0.377 & 0.365 & \textbf{0.422} & 0.089 & 0.067 \\
				\bottomrule
			\end{tabular}
		\end{table}
		
		\subsection{Comparison of heatmaps}
		
\begin{figure}[htb]
			\centering
			\subfloat[]{\includegraphics[width=0.48\textwidth]{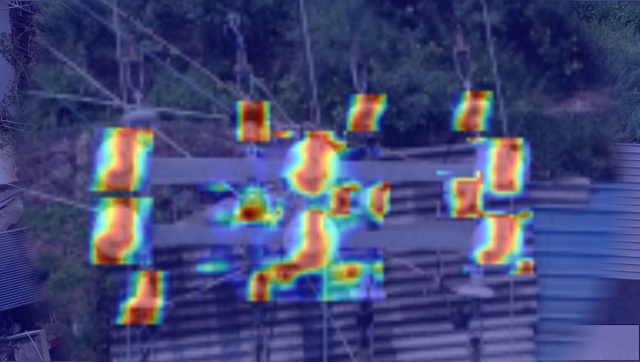}}\hfill
			\subfloat[]{\includegraphics[width=0.48\textwidth]{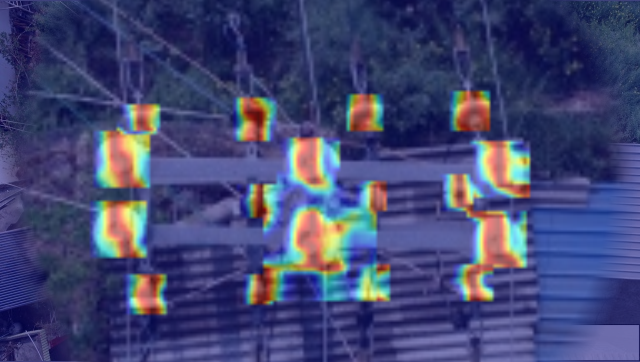}}\\
			
			\subfloat[]{\includegraphics[width=0.48\textwidth]{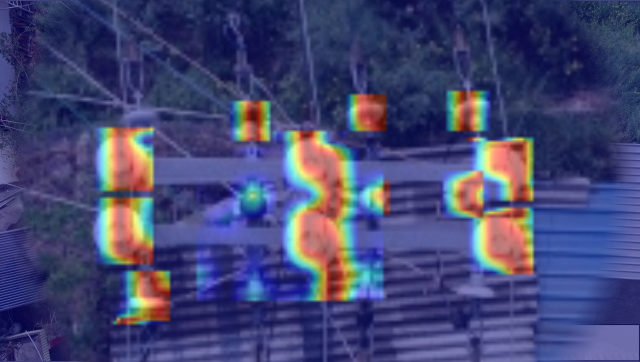}}\hfill
			\subfloat[]{\includegraphics[width=0.48\textwidth]{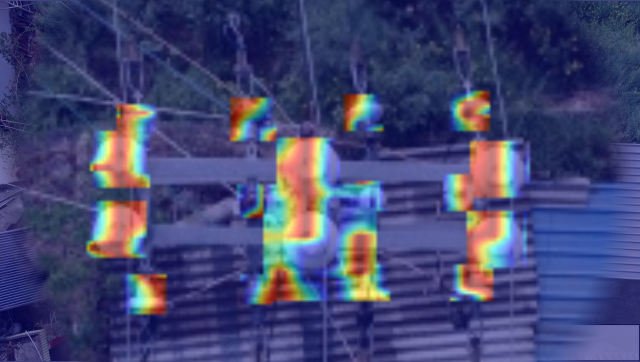}}\\
			
			\subfloat[]{\includegraphics[width=0.48\textwidth]{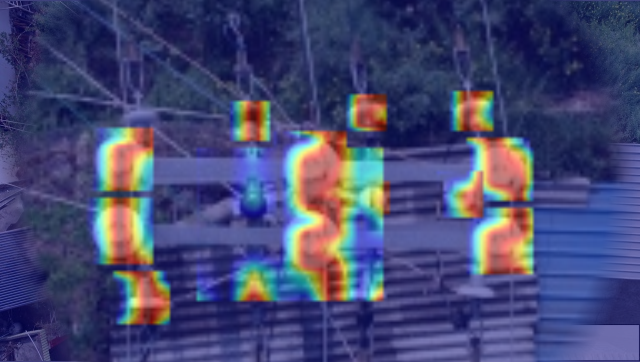}}\hfill
			\subfloat[]{\includegraphics[width=0.48\textwidth]{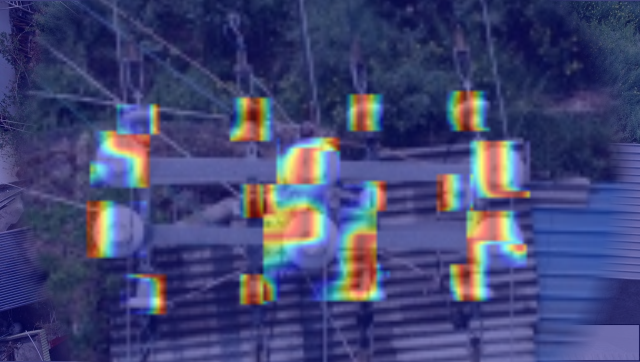}}
			
			\caption{Comparison of attention maps for the baseline RT-DETR model before and after incorporating different innovative modules: (a) without EEBlock, (b) with EEBlock, (c) without SPD Convolution, (d) with SPD Convolution, (e) without CSDMAM, (f) with CSDMAM.}
			\label{fig:example}
		\end{figure}

		To intuitively demonstrate the effectiveness of the proposed modules, we generate heatmaps from the baseline RT-DETR model and the model with the proposed modules (EE-ResNet, SPD Convolution, and CSDMAM) on the same input image. These heatmaps illustrate the attention distribution of the models during the recognition process. As shown in Fig.\ref{fig:example}, subfigures (a), (c), and (e) represent the attention maps from the baseline model corresponding to the modules before integration, while subfigures (b), (d), and (f) show the attention maps after incorporating the proposed modules. It can be observed that the attention distribution after the inclusion of the modules is more fine-grained, and the object contours are more distinct, indicating that the proposed modules enhance the model’s ability to preserve more detailed information. For better readability and to highlight the subtle differences, zoom-in views are provided, overlaid on the original images, allowing readers to clearly observe the enhanced attention patterns around key object regions.
		
		\subsection{Ablation Study of the Proposed Method}
		
		To analyze the impact of each component in CSG-ADCD, we conduct ablation studies by selectively enabling and disabling four key modules: EER (Edge-Enhanced ResNet), SPD (Stride-free Space-to-Depth), CSDMAM (Cross-Stage Dual-Domain Multi-Scale Attention Module), and FWS Loss (Focaler-Wise-SIoU Loss). The results in Table~\ref{tab:ablation_experiments} quantify each module’s contribution across multiple evaluation metrics. Here, \(\checkmark\) denotes an enabled module, and \(\times\) denotes a disabled one.
		
		\setcounter{table}{4}
		\begin{table}[htb]
			\centering
			\caption{Ablation Experiments for CSG-ADCD}
			\label{tab:ablation_experiments}
			\begin{tabular}{
					CCCC
					S[table-format=2.1]
					S[table-format=1.3]
					S[table-format=1.3]
					S[table-format=1.3]
					S[table-format=1.3]
				}
				\specialrule{0.8pt}{0pt}{1pt}
				EER & SPD & CSDMAM & FWS Loss &
				\multicolumn{1}{c}{GFLOPs} &
				\multicolumn{1}{c}{Precision} &
				\multicolumn{1}{c}{Recall} &
				\multicolumn{1}{c}{$\mathrm{mAP}_{50}$} &
				\multicolumn{1}{c}{$\mathrm{AP}_{s}$} \\
				\specialrule{0.5pt}{1pt}{0pt}
				\times & \times & \times & \times & 57.0 & 0.369 & 0.177 & 0.163 & 0.071 \\
				\checkmark & \times & \times & \times & 57.0 & 0.444 & 0.203 & 0.189 & 0.073 \\
				\times & \checkmark & \times & \times & 59.9 & 0.486 & 0.195 & 0.190 & 0.083 \\
				\times & \times & \checkmark & \times & \textbf{72.8} & 0.458 & 0.166 & 0.165 & 0.072 \\
				\times & \times & \times & \checkmark & 57.0 & 0.500 & 0.204 & 0.207 & 0.082 \\
				\checkmark & \checkmark & \times & \times & 59.7 & 0.417 & 0.208 & 0.192 & 0.076 \\
				\checkmark & \times & \checkmark & \times & 66.0 & 0.425 & 0.190 & 0.189 & 0.077 \\
				\times & \checkmark & \checkmark & \times & 65.2 & 0.385 & 0.192 & 0.182 & 0.072 \\
				\checkmark & \checkmark & \checkmark & \times & 65.3 & 0.485 & 0.233 & 0.237 & \textbf{0.119} \\
				\checkmark & \checkmark & \checkmark & \checkmark & 65.3 & \textbf{0.534} & \textbf{0.263} & \textbf{0.275} & 0.106 \\
				\specialrule{0.8pt}{0pt}{0pt}
			\end{tabular}
		\end{table}
		
		\textbf{Edge-Enhanced ResNet (EER)} EER strengthens boundary-aware features and reduces background leakage. Relative to the bare baseline, precision increases from 0.369 to 0.444, recall rises from 0.177 to 0.203, and $\mathrm{mAP}_{50}$ improves from 0.163 to 0.189 at an unchanged 57.0 GFLOPs. These shifts indicate fewer false positives at a fixed confidence threshold and more true positives on low-contrast contours. Importantly for the perceived instability, EER conditions the feature space so that subsequent attention behaves reliably. When EER is combined with SPD and CSDMAM, recall increases further to 0.233 and $\mathrm{AP}_{s}$ for small objects increases from 0.083 with SPD alone to 0.119 with all three. This pattern shows that EER anchors CSDMAM to genuine defect boundaries, turning a potentially volatile attention stage into a consistent contributor to small-object localization.
		
		\textbf{Space-to-Depth Convolution (SPD)} Applied alone, SPD preserves fine spatial evidence while avoiding aliasing, which suppresses spurious activations. Precision increases from 0.369 to 0.486, recall holds near 0.195, and $\mathrm{mAP}_{50}$ stays around 0.190, with GFLOPs rising modestly from 57.0 to 59.9. When SPD is paired with EER but without attention, recall increases from 0.195 to 0.208 and $\mathrm{mAP}_{50}$ nudges from 0.190 to 0.192, while precision softens from 0.486 to 0.417, which is expected because stronger edge cues surface more borderline candidates before a selective mechanism filters them. Once CSDMAM is added, those candidates are better gated: precision recovers from 0.417 to 0.485 and recall increases from 0.208 to 0.233. Thus, SPD and EER jointly deliver detail-rich, edge-faithful inputs that convert CSDMAM from a possible source of volatility into a consistent enhancer of small-object detection, as reflected by $\mathrm{AP}_{s}$ increasing from 0.083 to 0.119 when all three are enabled.
		
		\textbf{Cross-Stage Dual-Domain Multi-Scale Attention Module (CSDMAM)} The single-module ablation clarifies why a reviewer might perceive instability if attention is used in isolation. Enabling CSDMAM alone increases compute from 57.0 to 72.8 GFLOPs yet does not improve end-to-end accuracy: precision moves from 0.369 to 0.458, recall decreases from 0.177 to 0.166, and $\mathrm{mAP}_{50}$ remains near baseline at 0.165. This outcome indicates that, without edge priors or detail-preserving downsampling, attention can over-weight textured background regions and depress recall. Pairwise ablations corroborate this dependency: adding CSDMAM to SPD reduces precision from 0.486 to 0.385, and adding it to EER lowers precision from 0.444 to 0.425. In contrast, when CSDMAM is combined with both EER and SPD, its intended role materializes: $\mathrm{mAP}_{50}$ increases from 0.190 with SPD alone to 0.237 with all three modules, and $\mathrm{AP}_{s}$ for small objects increases from 0.083 to 0.119. Hence, CSDMAM is not intrinsically unstable; it becomes reliably beneficial once the backbone provides clean edges and SPD preserves the fine-scale evidence that attention needs to gate.
		
		\textbf{Focaler-Wise-SIoU Loss (FWS Loss)} FWS Loss converts representational gains into better calibration and tighter alignment. When applied to the EER–SPD–CSDMAM stack, precision increases from 0.485 to 0.534, recall increases from 0.233 to 0.263, and $\mathrm{mAP}_{50}$ increases from 0.237 to 0.275. These simultaneous gains show improved confidence shaping and stricter localization. Two interactions clarify the trade-offs. First, relative to the bare baseline, $\mathrm{AP}_{s}$ for small objects increases from 0.071 to 0.106 under the full configuration, confirming stronger tiny-target localization on average. Second, relative to the three-module model without the loss, $\mathrm{AP}_{s}$ decreases from 0.119 to 0.106 even as global metrics improve. This pattern reflects FWS Loss emphasizing medium-difficulty samples and penalizing loose boxes, which can prune marginal small-object detections at the chosen thresholds. In practice, this behavior is tunable by slightly relaxing the classification prior for the smallest anchors or moderating the modulation strength in the loss. Overall, FWS Loss interacts constructively with EER, SPD, and CSDMAM—stabilizing predictions, curbing over-confident false positives, and reinforcing the complementary effects of the modules—thereby addressing the instability observation within a co-designed pipeline.
		
		\subsection{Comparisons With Previous Methods}
		
		In this section, we conduct a rigorous comparative study between the proposed TinyDef-DETR and representative state-of-the-art detectors, including one-stage models (e.g., YOLOv5/8/10/11), two-stage/cascade detectors, and Transformer-based frameworks (e.g., RT-DETR, Conditional-DETR, DINO). All models are evaluated under a unified protocol—identical data splits, input resolutions, and training hyperparameters—to ensure a fair and reliable comparison.
		
\begin{figure}[H]
			\centering
			\subfloat[]{%
				\includegraphics[width=0.24\textwidth]{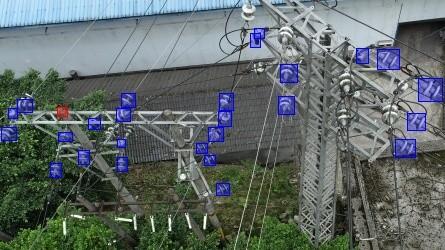}%
			}\hfill
			\subfloat[]{%
				\includegraphics[width=0.24\textwidth]{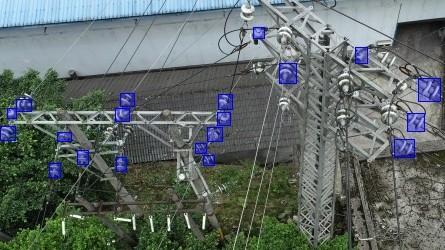}%
			}\hfill
			\subfloat[]{%
				\includegraphics[width=0.24\textwidth]{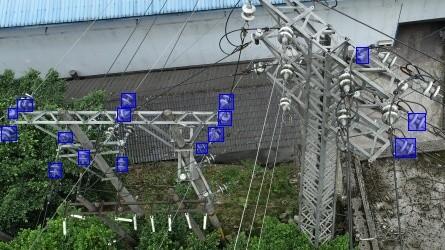}%
			}\hfill
			\subfloat[]{%
				\includegraphics[width=0.24\textwidth]{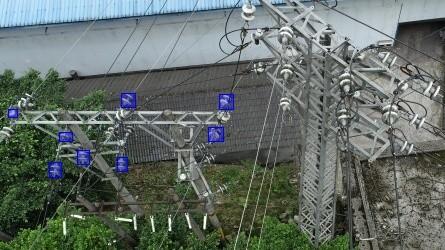}%
			}\\
			
			\subfloat[]{%
				\includegraphics[width=0.24\textwidth]{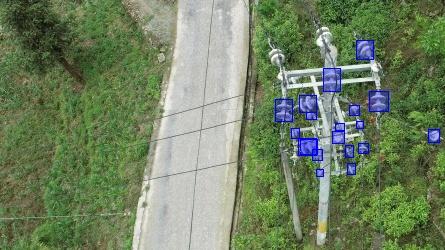}%
			}\hfill
			\subfloat[]{%
				\includegraphics[width=0.24\textwidth]{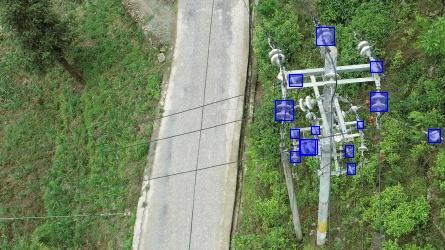}%
			}\hfill
			\subfloat[]{%
				\includegraphics[width=0.24\textwidth]{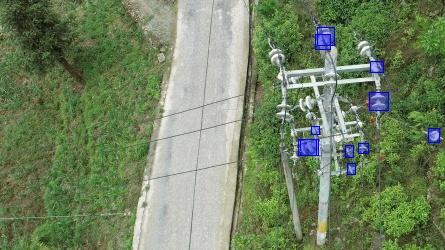}%
			}\hfill
			\subfloat[]{%
				\includegraphics[width=0.24\textwidth]{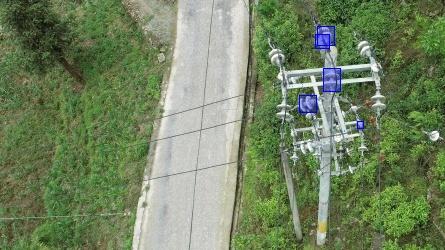}%
			}\\
			
			\subfloat[]{%
				\includegraphics[width=0.24\textwidth]{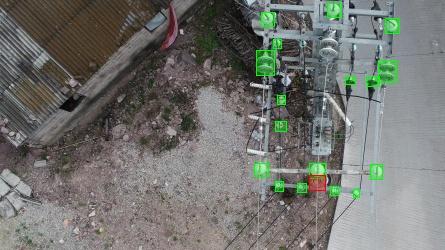}%
			}\hfill
			\subfloat[]{%
				\includegraphics[width=0.24\textwidth]{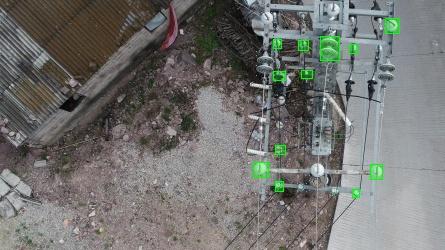}%
			}\hfill
			\subfloat[]{%
				\includegraphics[width=0.24\textwidth]{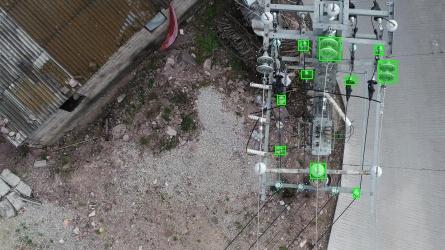}%
			}\hfill
			\subfloat[]{%
				\includegraphics[width=0.24\textwidth]{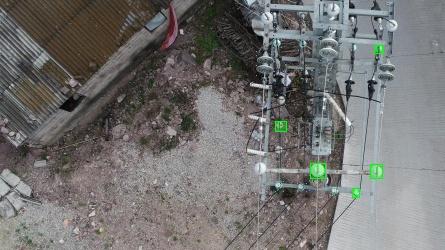}%
			}\\
			
			\subfloat[]{%
				\includegraphics[width=0.24\textwidth]{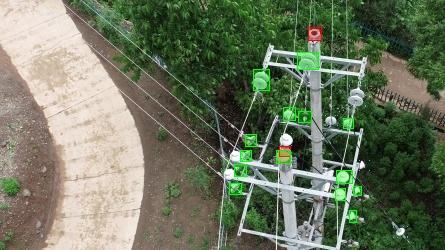}%
			}\hfill
			\subfloat[]{%
				\includegraphics[width=0.24\textwidth]{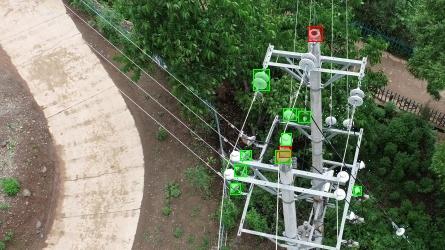}%
			}\hfill
			\subfloat[]{%
				\includegraphics[width=0.24\textwidth]{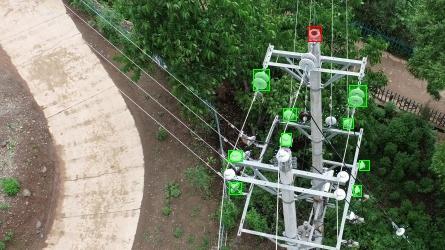}%
			}\hfill
			\subfloat[]{%
				\includegraphics[width=0.24\textwidth]{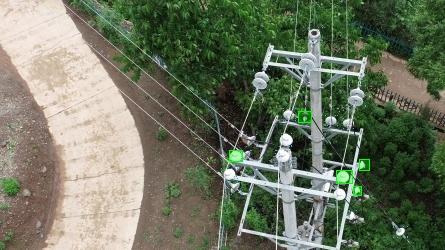}%
			}
			
			\caption{Comparison of detection results from four object detection methods on four representative test samples. Rows (a)–(d) correspond to different test images, and the four columns show the results obtained by TinyDef-DETR, DINO, RT-DETR-R18, and YOLO v5s, respectively.}
			\label{fig:results}
		\end{figure}

		Before presenting the quantitative results, we perform a qualitative case study on four representative images from the CSG-ADCD dataset to visually compare TinyDef-DETR with competing detectors. Since most targets in this dataset are small objects densely distributed in local regions, each image is cropped to focus on the areas with high object concentration, thereby providing a clearer and more informative visualization. To further enhance readability under such crowded conditions, detection results are rendered as filled regions rather than simple bounding box outlines. As illustrated in Fig.~\ref{fig:results}, TinyDef-DETR yields more accurate and consistent localization of small and densely packed objects compared with DINO, RT-DETR-R18, and YOLOv5s, clearly demonstrating its superiority in challenging small-object detection scenarios.
		
		Table~\ref{tab:model_performance} summarizes the overall performance of representative detectors on the CSG-ADCD dataset. The comparison covers lightweight and standard variants of YOLO (v5, v8, 10, 11), Transformer-based models such as RT-DETR and Conditional-DETR, as well as classical baselines including RetinaNet and DETR. Key indicators, including the number of parameters, computational complexity (GFLOPs), Precision, Recall, $\text{mAP}{50}$, and $\text{mAP}{50:95}$, are reported to provide a comprehensive view of accuracy–efficiency trade-offs. The best values in each column are highlighted in bold for clarity. Note that “YOLO 11s-P2” denotes a variant in which a P2 feature map was added to the model during training, and a dash (“-”) indicates that the corresponding measurement is not available.
		
		For a fair comparison, all methods are evaluated under a unified protocol—an input resolution of $640\times640$, identical NMS settings, and the same maximum number of detections per image.
		
		It is worth noting that DINO~\citep{zhangDINODETRImproved2022}—evaluated under \emph{identical} hardware and software configurations using the official implementation with its default 12-epoch schedule—performs worse than TinyDef-DETR across all reported metrics, while requiring 79.8 hours of training. In contrast, TinyDef-DETR, trained for 100 epochs under the same setup (following the canonical schedule adopted in RT-DETR\cite{zhaoDETRsBeatYOLOs2024}), completes in under 20 hours. Although wall-clock time is inherently environment-dependent, these numbers correspond to realistic, recommended settings. Importantly, the 12-epoch schedule for DINO and the 100-epoch schedule for TinyDef-DETR (inherited from RT-DETR) are the prescribed defaults; we therefore regard these schedules as intrinsic to each method and adopt them to compare truly “native” models. Under this protocol, TinyDef-DETR delivers a substantially more favorable accuracy–efficiency trade-off for real-world UAV inspection—amusingly, it seems this “longer training, better performance” effect doesn’t always make sense.
		
		\begin{table}[htb]
			\centering
			\caption{Model Performance Evaluation: Precision, Recall, and mAP for CSG-ADCD}
			\label{tab:model_performance}
			\setlength{\tabcolsep}{3pt} 
			\begin{tabular*}{\textwidth}{
					l
					@{\extracolsep{\fill}} 
					S[table-format=2.1]    
					S[table-format=3.1]    
					S[table-format=1.3]    
					S[table-format=1.3]    
					S[table-format=1.3, group-digits=false] 
					S[table-format=1.4, group-digits=false] 
				}
				\toprule
				Model & {Params (M)} & {GFLOPs} & {Precision} & {Recall} & {mAP$_{50}$} & {mAP$_{50:95}$} \\
				\midrule
				YOLO v5n & \textbf{2.2}  & \textbf{5.8}   & \textbf{0.707} & 0.131 & 0.131 & 0.0628 \\
				YOLO v5s & 9.1  & 23.8  & 0.346 & 0.148 & 0.140 & 0.0646 \\
				YOLO v5m & 25.1 & 64.0  & 0.548 & 0.164 & 0.162 & 0.0717 \\
				YOLO v8n & 2.7  & 6.8   & 0.580 & 0.134 & 0.133 & 0.0613 \\
				YOLO v8s & 11.1 & 28.5  & 0.375 & 0.155 & 0.147 & 0.0661 \\
				YOLO v8m & 23.2 & 67.5  & 0.403 & 0.169 & 0.163 & 0.0719 \\
				YOLO 10n\citep{wangYOLOv10RealtimeEndtoend2025} & 2.7  & 8.2   & \textbf{0.707} & 0.125 & 0.133 & 0.0627 \\
				YOLO 10s & 8.0  & 24.5  & 0.536 & 0.144 & 0.157 & 0.0725 \\
				YOLO 10m & 16.6 & 63.5  & 0.386 & 0.170 & 0.162 & 0.0753 \\
				YOLO 11n\citep{khanamYOLOv11OverviewKey2024a} & 2.6  & 6.3   & 0.638 & 0.135 & 0.136 & 0.0641 \\
				YOLO 11s & 9.4  & 21.3  & 0.516 & 0.166 & 0.155 & 0.0717 \\
				YOLO 11s-P2 & 9.6  & 28.7  & 0.608 & 0.183 & 0.181 & 0.0828 \\
				YOLO 11m & 20.0 & 67.7  & 0.399 & 0.181 & 0.173 & 0.0776 \\
				RT-DETR-l & 32.8 & 108.0 & 0.480 & 0.189 & 0.199 & 0.0890 \\
				RT-DETR-x & 65.5 & 222.5 & 0.485 & 0.199 & 0.216 & 0.0990 \\
				RT-DETR-R18\cite{zhaoDETRsBeatYOLOs2024} & 19.8 & 57.0  & 0.369 & 0.177 & 0.163 & 0.0738 \\
				RT-DETR-R34 & 31.2 & 88.8  & 0.354 & 0.127 & 0.117 & 0.0533 \\
				RT-DETR-R50 & 42.9 & 134.8 & 0.363 & 0.120 & 0.115 & 0.0514 \\
				RT-DETR-R101 & 74.7 & 247.1 & 0.467 & 0.185 & 0.193 & 0.0879 \\
				Conditional-DETR-R18\citep{mengConditionalDETRFast2023} & {--} & {--} & 0.309 & 0.068 & 0.038 & 0.0110 \\
				Conditional-DETR-R34 & {--} & {--} & 0.447 & 0.082 & 0.034 & 0.0090 \\
				Conditional-DETR-R50 & 44.0 & 89.5  & 0.568 & 0.159 & 0.172 & 0.0660 \\
				DETR-R50\citep{carionEndtoEndObjectDetection2020} & {--} & {--} & 0.301 & 0.138 & 0.124 & 0.0430 \\
				RetinaNet\citep{linFocalLossDense2018a} & 12.0 & 33.8  & 0.394 & 0.029 & 0.013 & 0.0040 \\
				DINO\citep{zhangDINODETRImproved2022} & {--} & {--} & 0.496 & 0.257 & 0.224 & 0.0986 \\
				FFCA-YOLO\citep{zhangFFCAYOLOSmallObject2024} & 7.1  & 51.4  & 0.680 & 0.130 & 0.142 & 0.0596 \\
				TinyDef-DETR & 20.5 & 65.3  & 0.534 & \textbf{0.263} & \textbf{0.275} & \textbf{0.1187} \\
				\bottomrule
			\end{tabular*}
		\end{table}
		
		\begin{table}[htb]
			\centering
			\caption{Model Performance Metrics: Recall, mAP$_{50}$, AP$_s$, and AP$_m$ for jyzwh and dgss}
			\label{tab:model_performance2}
			\setlength{\tabcolsep}{3pt} 
			\begin{tabular*}{\textwidth}{
					l
					@{\extracolsep{\fill}}
					S[table-format=1.3] 
					S[table-format=1.3] 
					S[table-format=1.3] 
					S[table-format=1.3] 
					S[table-format=1.3] 
				}
				\toprule
				Model & {Recall (jyzwh)} & {mAP$_{50}$ (jyzwh)} & {mAP$_{50}$ (dgss)} & {AP$_s$} & {AP$_m$} \\
				\midrule
				YOLO v5n & 0.352 & 0.315 & 0.258 & 0.061 & 0.097 \\
				YOLO v5s & 0.377 & 0.314 & 0.270 & 0.065 & 0.098 \\
				YOLO v5m & 0.372 & 0.367 & 0.277 & 0.079 & 0.081 \\
				YOLO v8n & 0.364 & 0.321 & 0.264 & 0.066 & 0.078 \\
				YOLO v8s & 0.388 & 0.331 & 0.269 & 0.067 & 0.095 \\
				YOLO v8m & 0.378 & 0.360 & 0.296 & 0.075 & 0.087 \\
				YOLO 10n\citep{wangYOLOv10RealtimeEndtoend2025} & 0.352 & 0.321 & 0.251 & 0.062 & 0.094 \\
				YOLO 10s & 0.343 & 0.341 & 0.308 & 0.078 & 0.095 \\
				YOLO 10m & 0.390 & 0.356 & 0.294 & 0.075 & 0.097 \\
				YOLO 11n\citep{khanamYOLOv11OverviewKey2024a} & 0.353 & 0.324 & 0.245 & 0.060 & 0.103 \\
				YOLO 11s & 0.381 & 0.344 & 0.291 & 0.074 & 0.100 \\
				YOLO 11s-P2 & 0.384 & 0.326 & 0.323 & 0.087 & 0.100 \\
				YOLO 11m & 0.382 & 0.373 & 0.307 & 0.082 & 0.094 \\
				RT-DETR-l & 0.360 & 0.341 & 0.372 & 0.078 & 0.044 \\
				RT-DETR-x & 0.366 & 0.363 & 0.403 & 0.083 & 0.058 \\
				RT-DETR-R18\cite{zhaoDETRsBeatYOLOs2024} & 0.347 & 0.316 & 0.354 & 0.071 & 0.039 \\
				RT-DETR-R34 & 0.293 & 0.260 & 0.284 & 0.056 & 0.062 \\
				RT-DETR-R50 & 0.278 & 0.254 & 0.275 & 0.054 & 0.027 \\
				RT-DETR-R101 & 0.345 & 0.316 & 0.378 & 0.073 & 0.050 \\
				Conditional-DETR-R18\citep{mengConditionalDETRFast2021} & 0.150 & 0.073 & 0.082 & 0.009 & 0.016 \\
				Conditional-DETR-R34 & 0.184 & 0.063 & 0.059 & 0.006 & 0.011 \\
				Conditional-DETR-R50 & \textbf{0.406} & 0.389 & 0.287 & 0.064 & 0.057 \\
				DETR-R50\citep{carionEndtoEndObjectDetection2020} & 0.289 & 0.280 & 0.203 & 0.041 & 0.037 \\
				RetinaNet\citep{linFocalLossDense2018a} & 0.048 & 0.022 & 0.006 & 0.012 & 0.003 \\
				FFCA-YOLO\citep{zhangFFCAYOLOSmallObject2024} & 0.286 & 0.291 & 0.270 & 0.059 & 0.051 \\
				TinyDef-DETR & 0.397 & \textbf{0.391} & \textbf{0.419} & \textbf{0.106} & \textbf{0.110} \\
				\bottomrule
			\end{tabular*}
		\end{table}
		
		\vspace{-8pt}
		
\begin{figure}[htb]
			\centering
			\includegraphics[width=0.9\textwidth]{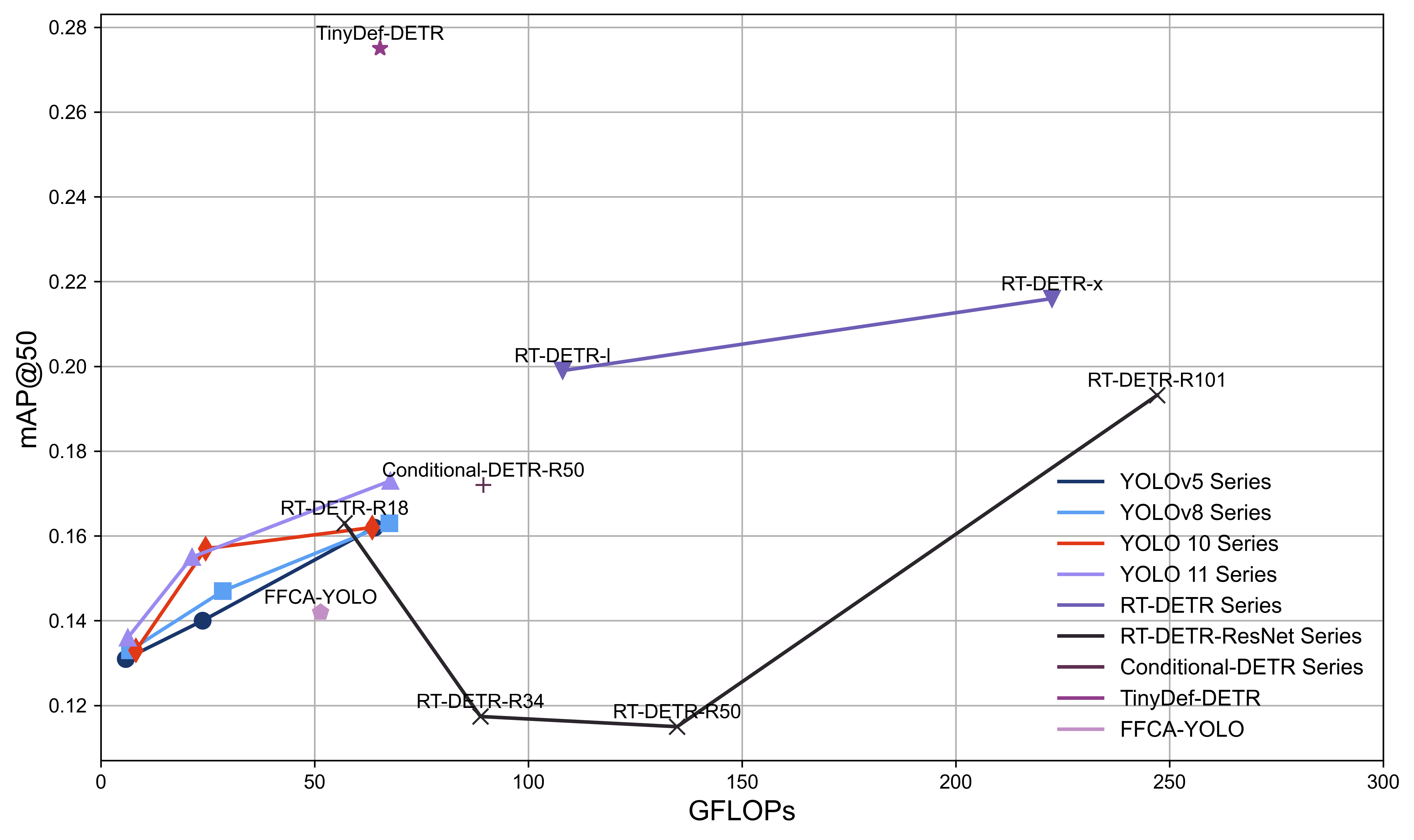}
			\caption{Comparison of model performance in terms of $\text{mAP}_{50}$ versus GFLOPs. 
				For clarity and to save space, labels for YOLO series are omitted; within each series, the models are ordered by GFLOPs from left to right as n, s, and m.}
			\label{fig:Comparison of Model Performance}
		\end{figure}

		A comprehensive illustration of the trade-off between detection accuracy and computational efficiency is presented in Fig.~\ref{fig:Comparison of Model Performance}, where the results reported in Table~\ref{tab:model_performance} are visualized. The plot highlights the relative positions of different detectors in terms of $\text{mAP}_{50}$ versus GFLOPs, enabling an intuitive comparison of the accuracy–efficiency balance across lightweight YOLO variants, Transformer-based detectors, and the proposed TinyDef-DETR. Notably, TinyDef-DETR achieves an $\text{mAP}_{50}$ of 0.275 with a computational cost of only 65.3 GFLOPs, striking a favorable balance between accuracy and efficiency compared with both lightweight YOLO models and heavier Transformer-based detectors. It is worth mentioning that while YOLO 11s-P2, which adds a P2 feature map to the original YOLO 11s, shows a moderate improvement in precision and recall (e.g., mAP$_{50}$ increases from 0.155 to 0.181), this simple addition of a P2 layer alone does not lead to a substantial boost in overall detection efficiency or accuracy–efficiency trade-off. This suggests that architectural modifications need to be more carefully designed to achieve meaningful gains, particularly for small-object detection in UAV inspection scenarios.
		
		Furthermore, Table~\ref{tab:model_performance2} provides a more detailed view of performance on small objects and defect-specific categories. It reports Recall and $\mathrm{mAP}_{50}$ for Polluted Insulator (jyzwh) and Damaged Pole (dgss) — the two most frequently occurring defect types in transmission line inspection — alongside scale-sensitive metrics $\mathrm{AP}_{s}$ and $\mathrm{AP}_{m}$. These results offer deeper insights into model behavior and enable fine-grained performance comparisons across representative defect categories. Although the dataset contains only 13 large objects, which is insufficient for robust evaluation, they are still part of transmission line defects. Our model achieves an $\mathrm{AP}_{l}$ of 0.454 for large objects, representing the average over IoU thresholds from 0.50 to 0.95, demonstrating its capability in detecting large-scale defects despite limited samples.
		
\begin{figure}[htb]
			\centering
			\includegraphics[width=0.8\textwidth]{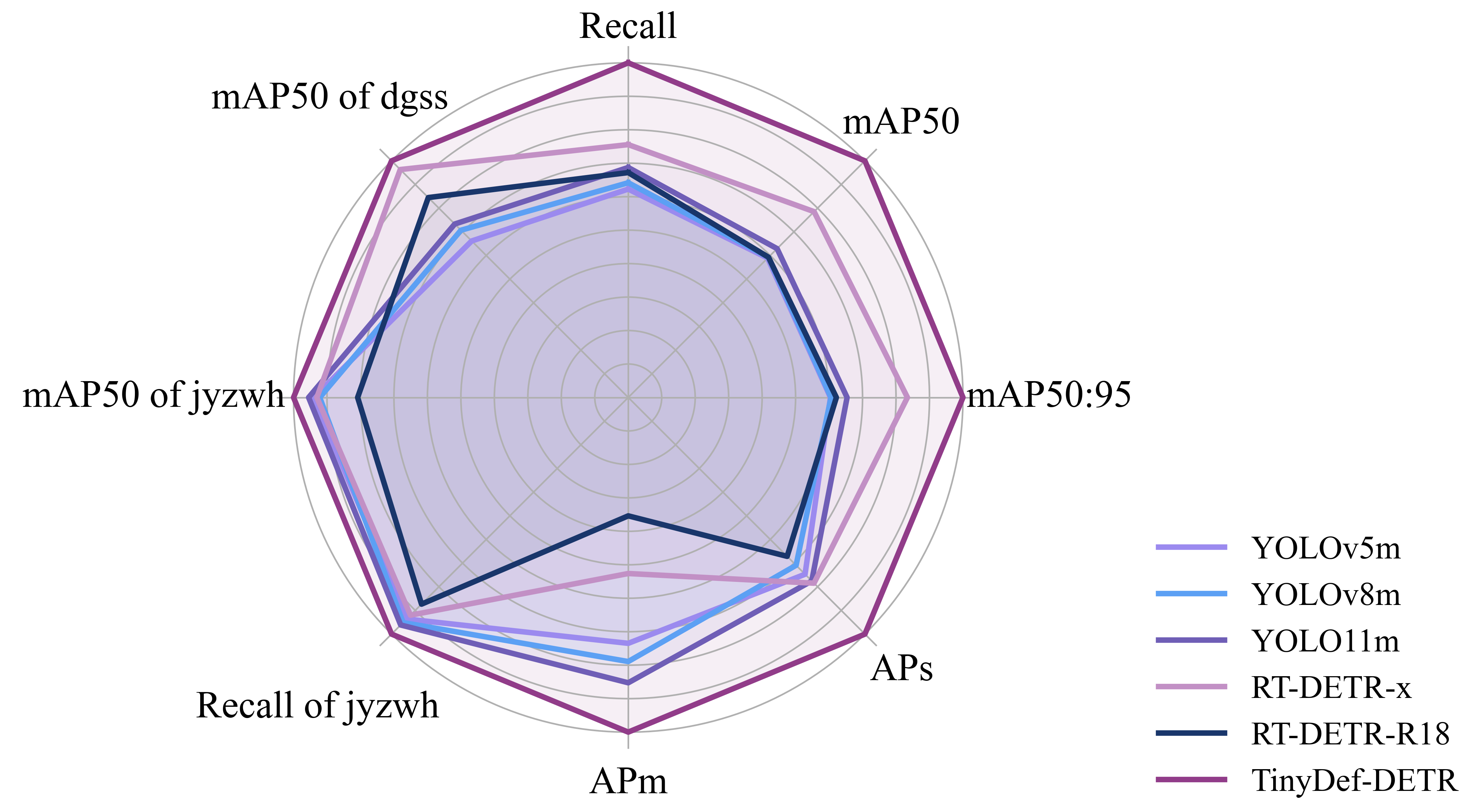}
			\caption{Radar Chart of Performance Comparison on Transmission Line Defect Detection}
			\label{fig:Radar}
		\end{figure}

		Finally, a holistic comparison of the selected detectors is presented in the form of a radar chart (Fig.~\ref{fig:Radar}). The models included — YOLO v5m, YOLO v8m, YOLO 11m, RT-DETR-R18, RT-DETR-x, and TinyDef-DETR — represent both one-stage and Transformer-based detection paradigms. The evaluation encompasses a comprehensive set of indicators, namely Recall, $\mathrm{mAP}_{50}$ , $\mathrm{mAP}_{50:95}$, $\mathrm{AP}_{s}$, $\mathrm{AP}_{m}$, Recall of Polluted Insulator (jyzwh), {mAP${50}$} of Polluted Insulator, and $\mathrm{mAP}_{50}$ of Damaged Pole (dgss). For enhanced visual comparability, each metric is normalized to its maximum value.
		
		\subsection{Generalization Experiments for the Proposed Methods}
		To further verify the generalization ability of the proposed method for small-object detection, we conduct experiments on VisDrone-DET 2019, a publicly available and generic UAV object detection benchmark rather than a domain-specific dataset. VisDrone-DET 2019~\cite{duVisDroneDET2019VisionMeets2019} contains 8{,}599 drone-captured images with over 540k annotated bounding boxes across ten general-purpose categories (e.g., pedestrian, person, car, van, bus, truck, motor, bicycle, awning-tricycle, tricycle), split into 6{,}471/548/1{,}580 for train/val/test. Crucially, VisDrone does not include transmission-line defect classes; thus, our results on VisDrone (Table~\ref{tab:comparison_vis}) explicitly measure cross-domain transferability rather than in-domain recognition. Owing to its dense object distributions, the high prevalence of small targets, and complex real-world scenes, VisDrone offers a rigorous testbed for evaluating robustness and generalization.
		
		As shown in Table~\ref{tab:comparison_vis}, TinyDef-DETR attains leading overall performance with moderate model complexity. Concretely, it consistently outperforms strong baselines across standard detection metrics and shows clear advantages on small objects, while keeping computation and parameters at a practical level.
		
		These results substantiate that the proposed framework generalizes effectively to small-object detection in UAV imagery beyond transmission-line inspection, indicating practical applicability across domains with different object semantics.
		
		\begin{table}[htb]
			\centering
			\caption{Comparison Experiments for TinyDef-DETR on VisDrone}
			\label{tab:comparison_vis}
			
			\begin{tabular*}{\linewidth}{@{\hspace{\tabcolsep}}
					l @{\extracolsep{\fill}}
					S[table-format=2.1]
					S[table-format=3.1]
					S[table-format=1.3]
					S[table-format=1.3]
					S[table-format=1.3]
					S[table-format=1.3]
				}
				\specialrule{1pt}{0pt}{1pt}
				{Model} & {Params (M)} & {GFLOPs} & {mAP\(_{50}\)} & {mAP\(_{50:95}\)} & {AP\(_s\)} & {AP\(_m\)} \\
				\specialrule{0.6pt}{1pt}{1pt}
				YOLO v8n & 3.0  & 8.1  & 0.259 & 0.144 & 0.059 & 0.225 \\
				YOLO v8s & 11.1 & 28.5 & 0.307 & 0.173 & 0.078 & 0.269 \\
				YOLO v8m & 25.9 & 78.7 & 0.332 & 0.190 & 0.090 & 0.294 \\
				YOLO 10n\citep{wangYOLOv10RealtimeEndtoend2025} & 2.3  & 6.5  & 0.261 & 0.142 & 0.063 & 0.224 \\
				YOLO 10s & 7.2  & 21.4 & 0.323 & 0.179 & 0.086 & 0.278 \\
				YOLO 10m & 15.3 & 58.9 & 0.345 & 0.195 & 0.097 & 0.300 \\
				YOLO 11n\citep{khanamYOLOv11OverviewKey2024a} & 2.6  & 6.3  & 0.258 & 0.142 & 0.058 & 0.225 \\
				YOLO 11s & 9.4  & 21.3 & 0.313 & 0.176 & 0.080 & 0.272 \\
				YOLO 11m & 20.0 & 67.7 & 0.350 & 0.203 & 0.098 & 0.312 \\
				YOLO 12n\citep{tianYOLOv12AttentionCentricRealTime2025} & 2.6  & 6.3  & 0.259 & 0.142 & 0.057 & 0.224 \\
				YOLO 12s & 9.2  & 21.2 & 0.312 & 0.176 & 0.081 & 0.274 \\
				YOLO 12m & 20.1 & 67.2 & 0.336 & 0.192 & 0.094 & 0.298 \\
				YOLO 13n\citep{leiYOLOv13RealTimeObject2025} & 2.5  & 6.2  & 0.244 & 0.133 & 0.055 & 0.210 \\
				YOLO 13s & 9.0  & 20.1 & 0.297 & 0.167 & 0.077 & 0.258 \\
				YOLOX-Tiny\citep{geYOLOXExceedingYOLO2021} & 5.0  & 7.6  & 0.278 & 0.148 & 0.076 & 0.221 \\
				RT-DETR-R18\cite{zhaoDETRsBeatYOLOs2024} & 20.0 & 60.0 & 0.333 & 0.185 & 0.139 & 0.275 \\
				ATSS-R50\citep{zhangBridgingGapAnchorbased2020a} & 38.9 & 110.0 & 0.338 & 0.204 & 0.100 & 0.317 \\
				Cascade-RCNN-R50\citep{caiCascadeRCNNHigh2021} & 69.3 & 236.0 & 0.326 & 0.197 & 0.099 & 0.309 \\
				D-Fine-N\citep{pengDFINERedefineRegression2024} & 3.7  & 7.1  & 0.334 & 0.183 & 0.093 & 0.270 \\
				Faster-RCNN-R50\citep{renFasterRCNNRealTime2015a} & 41.4 & 208.0 & 0.329 & 0.194 & 0.095 & 0.309 \\
				FBRT-YOLO-M\citep{xiaoFBRTYOLOFasterBetter} & 7.4  & 58.7 & 0.344 & 0.196 & 0.094 & 0.309 \\
				FBRT-YOLO-N & 0.8  & 6.7  & 0.265 & 0.148 & 0.062 & 0.234 \\
				FBRT-YOLO-S & 2.9  & 22.9 & 0.323 & 0.183 & 0.085 & 0.283 \\
				FFCA-YOLO\citep{zhangFFCAYOLOSmallObject2024} & 7.1  & 51.4 & 0.267 & 0.122 & 0.091 & 0.295 \\
				RetinaNet-R50\citep{linFocalLossDense2018a} & 36.5 & 210.0 & 0.276 & 0.164 & 0.060 & 0.274 \\
				RTMDet-Tiny\citep{lyuRTMDetEmpiricalStudy2022} & 4.9  & 8.0  & 0.312 & 0.184 & 0.077 & 0.288 \\
				TOOD-R50\citep{fengTOODTaskalignedOnestage2021} & 32.0 & 199.0 & 0.339 & 0.204 & 0.102 & 0.317 \\
				TinyDef-DETR & 20.5 & 65.0 & {\bfseries 0.372} & {\bfseries 0.217} & {\bfseries 0.148} & {\bfseries 0.348} \\
				\specialrule{1pt}{1pt}{0pt}
			\end{tabular*}
		\end{table}
		\section{Discussion}
		
		\subsection{Theoretical Mechanisms of Performance Enhancement}
		
		The performance gains observed with TinyDef-DETR arise from the mutually reinforcing effects of its architectural and loss-design innovations, each addressing a specific bottleneck in UAV-based defect detection. At the representational level, the edge-enhanced ResNet backbone strengthens the model’s sensitivity to fine structural cues by amplifying high-frequency components associated with defect boundaries. This modification counteracts the intrinsic attenuation of edge information caused by deep convolutional downsampling, enabling the model to preserve discriminative contours essential for identifying small or visually ambiguous defects. From a theoretical standpoint, such enhancement effectively increases the signal-to-noise ratio for fine-grained features and improves the separability of defect patterns within the feature manifold.
		
		The stride-free space-to-depth (SPD) module introduces a second source of improvement by performing spatial reduction without discarding pixel-level detail. Unlike conventional stride-based downsampling, which induces aliasing and information loss, SPD reorganizes high-resolution inputs into channel-enriched tensors. This preserves local texture, minimizes feature degradation for small targets, and stabilizes gradient flow. Theoretically, this operation enhances the mutual information between shallow and deep features, enabling a more coherent propagation of small-object descriptors throughout the network hierarchy.
		
		Furthermore, the cross-stage dual-domain multi-scale attention mechanism integrates spatial and frequency-domain representations to model global context and local variations more holistically. By enabling multi-scale interactions across network stages, this module addresses the scale variability characteristic of UAV imagery and improves the model’s ability to capture long-range structural dependencies. In essence, the mechanism expands the effective receptive field while simultaneously preserving the locality required for small-object detection, creating a balanced and robust feature fusion process.
		
		Finally, the Focaler--Wise--SIoU loss contributes an optimization-level enhancement by modulating regression difficulty through an EMA-normalized difficulty indicator. The unimodal weighting function prioritizes moderately hard samples while attenuating extreme outliers and trivial cases, resulting in a smoother and more stable error landscape. This adaptive reweighting improves convergence behavior, enhances bounding-box localization for small defects, and mitigates gradient domination by noisy samples. From a theoretical perspective, the loss introduces a dynamic sample-importance prior that aligns training emphasis with statistically informative examples, thereby improving both robustness and generalization.
		
		Collectively, these mechanisms form a coherent theoretical foundation explaining why TinyDef-DETR exhibits strong performance on small, ambiguous, and cluttered defect instances in UAV imagery. Each component contributes a targeted enhancement---representational fidelity, information preservation, multi-scale context integration, and difficulty-aware optimization---that together yield a substantial improvement over conventional detection pipelines.
		
		\subsection{Limitations and Future Directions}
		
		Despite its effectiveness, TinyDef-DETR still exhibits certain limitations that warrant further exploration. First, although SPD alleviates information loss during downsampling, the increased channel dimensionality demands additional memory and may introduce latency on resource-limited embedded UAV platforms. Lightweight kernel decomposition or dynamic channel pruning may help reduce overhead while preserving detail fidelity.
		
		Second, the dual-domain attention module, while beneficial for multi-scale fusion, introduces additional complexity to the training pipeline. In scenarios involving extremely high-resolution imagery or large-scale datasets, the global context modeling may become computationally expensive. Future work may explore hierarchical or sparse attention variants to reduce computational burden while maintaining cross-stage consistency.
		
		Third, the difficulty-modulated regression loss relies on reliable EMA statistics of IoU distributions. In datasets with extreme imbalance or rapidly shifting difficulty patterns, the moving average may adapt too slowly, reducing the responsiveness of the modulation mechanism. Developing more flexible or instance-adaptive normalization strategies could further stabilize optimization and improve performance on highly heterogeneous data.
		
		Additionally, although the present study focuses on transmission line defect detection, the generalization of TinyDef-DETR to other aerial inspection domains---such as railway, bridge, and pipeline monitoring---remains to be systematically evaluated. Cross-domain validation, domain adaptation methods, or self-supervised pretraining may help broaden its applicability.
		
		Finally, the current framework does not explicitly incorporate geometric priors, structural constraints, or transformer-based depth modeling that could further improve detection under severe occlusion or low-contrast imaging. Integrating 3D contextual cues, physical-structure modeling, or multi-modal inputs (e.g., infrared or LiDAR) could provide additional robustness in complex operational environments.	
		\section{Conclusions}
		
		In this paper, we introduced TinyDef-DETR, a DETR-based framework tailored to defect detection in UAV imagery of transmission lines. The design combines an Edge-Enhanced ResNet backbone, a stride-free space-to-depth (SPD) downsampling module, a cross-stage dual-domain multi-scale attention mechanism, and a Focaler–Wise–SIoU regression loss. Together, these components target the core difficulties of this task—tiny object scale, visual ambiguity, and background clutter—by strengthening edge/detail representation, preserving pixel-level cues during downsampling, enhancing global–local feature interaction, and stabilizing box regression for small targets.
		
		Comprehensive experiments on the CSG-ADCD dataset and real-world UAV imagery show that TinyDef-DETR consistently surpasses representative detectors in detection accuracy, recall, and robustness, while maintaining competitive computational efficiency. Ablation studies further confirm that each module contributes complementary gains to the final system, clarifying the role of edge-aware features, lossless downsampling, dual-domain attention, and shape/angle-aware localization in the overall performance.
		
		These results indicate that TinyDef-DETR delivers strong detection capability for small and challenging defects and is well-suited for practical UAV-based inspection. Owing to its modular and lightweight design, the framework is readily adaptable to broader power-system inspection workflows and can be extended to other small-object detection problems in aerial and remote-sensing applications.
		
		\vspace{6pt} 
		
		
		
		
		
		\authorcontributions{Conceptualization, J.C. and S.Z.; methodology, J.C.; software, J.C.; validation, J.C. and F.S.; formal analysis, J.C.; investigation, J.C. and W.L.; resources, F.S. and S.Z.; data curation, J.C.; writing—original draft preparation, J.C.; writing—review and editing, S.Z. and F.S.; visualization, J.C.; supervision, S.Z.; project administration, S.Z.; funding acquisition, S.Z. All authors have read and agreed to the published version of the manuscript.}
		
		\funding{This work was supported by the National Natural Science Foundation of China under Grant 61673128, Grant 61573117, and Grant 41627801. (Corresponding author: Feng Shen.)}
		
		\institutionalreview{Not applicable.}
		
		\informedconsent{Not applicable.}
		
		\dataavailability{Due to confidentiality agreements, part of the dataset used in this study cannot be shared publicly. However, the data presented in this study also include openly available resources from the VisDrone benchmark dataset, which can be accessed at \url{https://github.com/VisDrone/VisDrone-Dataset}, as described in \cite{duVisDroneDET2019VisionMeets2019}.}

		\acknowledgments{The authors would like to thank the organizers of the VisDrone dataset for providing the publicly available data that supported this research. The authors also thank the developers of RT-DETR for their open-source contributions, which greatly facilitated the experiments in this work.}
		
		\conflictsofinterest{The authors declare no conflicts of interest.} 
		
			\begin{adjustwidth}{-\extralength}{0cm}
			
			\reftitle{References}
			
			
			\bibliography{reference.bib}{}

			%
			
			
			\PublishersNote{}
			\end{adjustwidth}
		
	\end{document}